\documentclass[10pt,twocolumn,letterpaper]{article}

\usepackage{cvpr}              % To produce the CAMERA-READY version
\usepackage[dvipsnames]{xcolor}

\definecolor{cvprblue}{rgb}{0.21,0.49,0.74}
\usepackage[pagebackref,breaklinks,colorlinks,citecolor=cvprblue]{hyperref}

\usepackage[capitalize]{cleveref}
\crefname{section}{Sec.}{Secs.}
\Crefname{section}{Section}{Sections}
\Crefname{table}{Table}{Tables}
\crefname{table}{Tab.}{Tabs.}
\usepackage{amssymb}
\usepackage{wrapfig}
\usepackage{lipsum}
\usepackage{setspace}
\usepackage{multirow}
\definecolor{lightgray}{gray}{0.9} 
\usepackage{colortbl}
\usepackage{xcolor}
\usepackage{times}
\usepackage[accsupp]{axessibility} % Improves PDF readability for those with visual impairments.
\newcommand{\myparagraphZ}[1]{\vspace{7pt}\noindent{\bf #1}\hspace{7pt}}

\def\paperID{16074} % *** Enter the Paper ID here
\def\confName{CVPR}
\def\confYear{2024}

\title{Open Vocabulary Semantic Scene Sketch Understanding}

\author{Ahmed Bourouis\textsuperscript{1}
\and 
Judith E. Fan\textsuperscript{2}
\and 
Yulia Gryaditskaya\textsuperscript{1}
\and\\
\textsuperscript{1}{Surrey Institute for People-Centered AI, CVSSP, University of Surrey, UK}\\
\textsuperscript{2}{Department of Psychology, Stanford University, USA}\\
{\large\textcolor{gray}{\href{https://ahmedbourouis.github.io/Scene_Sketch_Segmentation/}{https://ahmedbourouis.github.io/Scene\_Sketch\_Segmentation/}}} % Smaller font and 50% opacity
} 

\begin{document}
\maketitle

\begin{abstract}
% We study the underexplored but fundamental vision problem of machine understanding of abstract freehand scene sketches.
% We introduce a sketch encoder that results in semantically-aware feature space, which we evaluate by testing its performance on a semantic sketch segmentation task.
% To train our model we rely only on the availability of bitmap sketches with their brief captions and do not require any pixel-level annotations. 
% To obtain generalization to a large set of sketches and categories, we build on a vision transformer encoder pretrained with the CLIP model. 
% We freeze the text encoder and perform visual-prompt tuning of the visual encoder branch while introducing a set of critical modifications. 
% Firstly, we augment the classical key-query (k-q) self-attention blocks with value-value (v-v) self-attention blocks. 
% Central to our model is a two-level hierarchical training that enables efficient semantic disentanglement: The first level ensures holistic scene sketch encoding, and the second level focuses on individual categories.
% We, then, in the second level of the hierarchy, introduce a cross-attention between textual and visual branches.
% Our method outperforms zero-shot CLIP 
% % pixel accuracy of 
% segmentation results by 37 points, reaching a pixel accuracy of $85.5\%$ on the FS-COCO sketch dataset.
% Finally, we conduct a user study that allows us to identify further improvements needed over our method to reconcile machine and human understanding of freehand scene sketches.

We study the underexplored but fundamental problem of machine understanding of abstract freehand scene sketches.
We introduce a sketch encoder that ensures a semantically-aware feature space, which we evaluate by testing its performance on a semantic sketch segmentation task.
To train our model, we rely only on bitmap sketches accompanied by brief captions, avoiding the need for pixel-level annotations. 
To generalize to a large set of sketches and categories, we build upon a vision transformer encoder pretrained with the CLIP model. 
We freeze the text encoder and perform visual-prompt tuning of the visual encoder branch while introducing a set of critical modifications. 
First, we augment the classical key-query (k-q) self-attention blocks with value-value (v-v) self-attention blocks. 
Central to our model is a two-level hierarchical training that enables efficient semantic disentanglement: The first level ensures holistic scene sketch encoding, and the second level focuses on individual categories.
In the second level of the hierarchy, we introduce cross-attention between the text and vision branches.
Our method outperforms zero-shot CLIP segmentation results by 37 points, reaching a pixel accuracy of $85.5\%$ on the FS-COCO sketch dataset.
Finally, we conduct a user study that allows us to identify further improvements needed over our method to reconcile machine and human understanding of freehand scene sketches.

% Code is available at \href{https://github.com/AhmedBourouis/Scene-Sketch-Segmentation}{https://github.com/AhmedBourouis/Scene-Sketch-Segmentation}.
% Project page \href{https://ahmedbourouis.github.io/Scene_Sketch_Segmentation/}{https://ahmedbourouis.github.io/Scene\_Sketch\_Segmentation/}
\end{abstract}

% \vspace{-0.6cm}

% We study the underexplored but fundamental problem of machine understanding of freehand scene sketches.
% We introduce a sketch encoder that acquires a semantically-aware feature space, which we evaluate by testing its performance on a semantic sketch segmentation task.
% To train our model we rely only on bitmap sketches accompanied by brief captions, avoiding the need for pixel-level annotations. 
% To generalize to a large set of sketches and categories, we build upon a CLIP-based vision transformer encoder. 
% We freeze the text encoder and perform visual-prompt tuning of the visual encoder branch while introducing a set of critical modifications. 
% First, we augment the classical key-query (k-q) self-attention blocks with value-value (v-v) self-attention blocks. 
% Central to our model is a two-level hierarchical training that enables efficient semantic disentanglement: The first level ensures holistic scene sketch encoding, and the second level focuses on individual categories.
% In the second level of the hierarchy, we introduce cross-attention between the text and vision branches.
% Our method outperforms zero-shot CLIP segmentation results by 37 points, reaching a pixel accuracy of $85.5\%$ on the FS-COCO sketch dataset.
% Finally, we conduct a user study that allows us to identify further improvements needed over our method to reconcile machine and human understanding of freehand scene sketches.    

\section{Introduction}
\label{sec:intro}

Even a quick sketch can convey rich information about what is relevant in a visual scene: what objects there are and how they are arranged. 
However, little work has been devoted to the task of machine scene sketch understanding, largely due to a lack of data. 
Understanding sketches with methods designed for images is challenging because sketches have very different statistics from images -- they are sparser and lack detailed color and texture information. 
Moreover, sketches contain abstraction at multiple levels: the holistic scene level and the object level.
Here we explore the promise of two main ideas: (1) the use of language to guide the learning of how to parse scene sketches and (2) a two-level training network design for holistic scene understanding and individual categories recognition.

\begin{figure}[t]
  \centering
  %\fbox{\rule{0pt}{2in} \rule{0.9\linewidth}{0pt}}
   \includegraphics[width=1.0\linewidth]{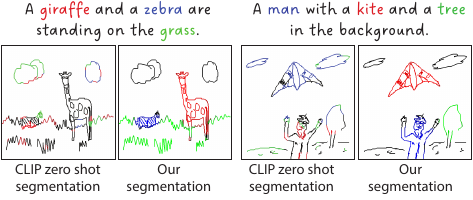}
    \vspace{-0.7cm}
   \caption{Comparison of the segmentation result obtained with CLIP visual encoder features and features from our model.}
   \label{fig:teaser}
     \vspace{-0.4cm}
\end{figure}

\begin{figure*}
  \centering
    \includegraphics[width=\linewidth]{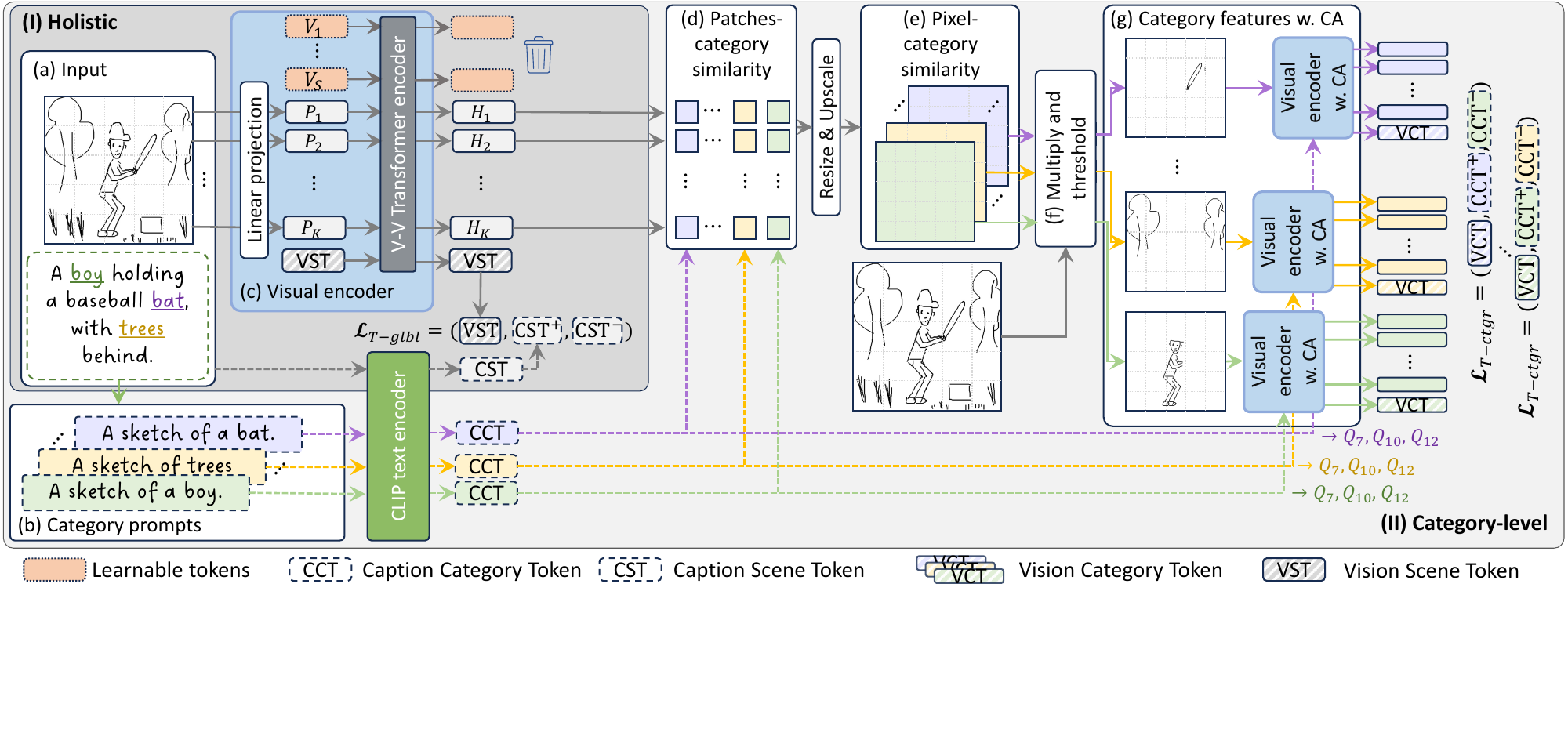}
\vspace{-2.1cm}
   \caption{Our framework consists of two levels: I. Holistic Scene Sketch Understanding and II. Targeting individual categories disentanglement. Please refer to \cref{sec:method} for details.}
   \label{fig:pipeline}
\end{figure*}

Freehand sketches can be represented as a sequence or cloud of individual strokes, or as a bitmap image. 
As one of the first works on scene sketch understanding, we target a general setting where we assume only the availability of bitmap representations. 
We also aim at the method that can generalize to a large number of scenes and object categories. 
To this end, we build our sketch encoder on a Visual Transformer (ViT) encoder pre-trained with a popular CLIP \cite{radford2021learning} foundation model (\cref{fig:teaser}). 
We propose a two-level hierarchical training of our network, where the two levels (``Holistic" and ``Category-level") share the weights of our visual encoder. 
The first level focuses on ensuring that our model can capture holistic scene understanding (\cref{fig:pipeline}: I. Holistic), while the second level ensures that the encoder can efficiently encode and distinguish individual categories (\cref{fig:pipeline}: II. Category-level). 
We avoid reliance on tedious user per-pixel annotations by leveraging sketch-caption pairs from the FS-COCO dataset \cite{chowdhury2022fs}, and aligning the visual tokens of sketch patches with textual tokens from the sketch captions, using triplet loss training. 
We strengthen the alignment by introducing sketch-text cross-attention in the second level of the network's hierarchy (\cref{fig:pipeline}: g.). 
Additionally, we introduce a modified self-attention computation to the visual transformer encoder used in both layers, inspired by recent work by Li et al.~\cite{li2023clip}. 

We conduct a comprehensive evaluation of our method comparing it with recent language-supervised image segmentation methods \cite{ xu2022groupvit,radford2021learning, li2023clip}, fine-tuned on the FS-COCO dataset. 
We show that our approach outperforms with a large margin all existing methods on the task of freehand sketch segmentation. 
We also compare with a previous fully supervised work on scene sketch segmentation \cite{ge2022exploring}, trained on a semi-synthetic set of sketches composed of individual category sketches.
We demonstrate that their work does not generalize well to freehand scene sketches \cite{chowdhury2022fs}. 
% from the FS-COCO dataset.
Our method demonstrates consistent performance and similarly outperforms \cite{ge2022exploring} on a dataset of freehand sketches provided by Ge \etal \cite{ge2022exploring}.

Finally, our analysis reveals that although our model consistently produces robust segmentation results across the majority of sketches, there are a few challenging sketching scenarios for our method. 
We select a subset of representative sketches for each scenario and collect multi-user annotations.
We then carefully assess our approach by comparing its performance with that of human participants, drawing insights to guide future work.

In summary, our contributions include:
(1) a two-level hierarchical training approach, focusing on holistic scene sketch understanding and category disentanglement,
(2) the first language-supervised scene sketch segmentation method,
(3) per pixel segmentation annotations of 975 sketches from the FS-COCO dataset,
and
(4) multi-user annotations of a subset of distinct groups of sketches. 

\section{Related Work}
\label{sec: related works}
\subsection{Unsupervised and Weekly Supervised Image Semantic Segmentation}
% \paragraph{Unsupervised and weekly supervised image semantic segmentation}
The need for pixel-wise segmentation limits the number of instances that supervised segmentation models \cite{long2015fully,zhao2017pyramid,chen2017deeplab,chen2017rethinking,badrinarayanan2017segnet,fu2019dual} can use for training, as such annotations are costly to collect.
This in turn limits the generalization properties of models trained with pixel-level annotations. 
To avoid the need for extensive annotations, unsupervised \cite{zadaianchuk2022unsupervised,cho2021picie,hamilton2022unsupervised,hwang2019segsort,melas2022deep}, semi-supervised \cite{mittal2019semi,zhu2021improving} and weakly supervised \cite{pathak2014fully,wei2018revisiting,xu2022groupvit,luo2022segclip,dong2023maskclip,ding2022decoupling,luddecke2022image,zhou2023zegclip,he2023clips4} methods were proposed.

Our method belongs to the group of weakly supervised methods based on text annotations only \cite{xu2022groupvit,luo2022segclip,dong2023maskclip,ding2022decoupling,luddecke2022image,chen2023exploring}, such methods are not limited to a fixed set of categories and therefore are referred to as open vocabulary semantic segmentation methods. 
Image methods typically rely on the spatial proximity of semantically similar pixels.
This is less applicable in the sparse and largely monochromatic landscape of freehand sketches.
For example, recent GroupViT \cite{xu2022groupvit} and SegCLIP \cite{luo2022segclip} use learnable group tokens and semantic group modules to aggregate low-layer pixel features. 
In our work, we propose a two-level training architecture taking sketch sparsity and abstraction into account.

\subsection{Sketch Semantic Segmentation} 
% \paragraph{Sketch semantic segmentation} 
The majority of works on semantic sketch segmentation focus on single-category sketches.
Some of these works treat sketch as a bitmap image \cite{zhu2018part, li2018fast, zhu20202d}, but most leverage stroke-level information directly \cite{ha2017neural,kaiyrbekov2019deep,wu2018sketchsegnet,qi2019sketchsegnet+,defferrard2016convolutional,scarselli2008graph,hahnlein2019bitmap, wang2020multi, yang2020sketchgcn,qi2022one,zheng2023sketch} or as a segmentation refinement step \cite{li2018fast,zhu20202d}. 
All these works are fully supervised except for \cite{qi2022one}, which segments sketches of a given category provided at least one segmented reference sketch. 

Semantic scene sketch segmentation \cite{sun2012free}, and more broadly scene sketch understanding, is underexplored, to a large extent due to a lack of data. 
% Semantic scene sketch segmentation \cite{sun2012free}, as an instance of sketch understanding task, is under-explored, to a large extent due to a lack of data. 
The lack of data is typically addressed by introducing semi-synthetic sketch datasets.
The SketchyScene dataset \cite{zou2018sketchyscene} consists of 7,264 sketch-image pairs, obtained by arranging clip-art individual category sketches in alignment with a reference image. 
SketchyCOCO dataset \cite{gao2020sketchycoco} is generated from COCO-Stuff \cite{caesar2018coco} by semi-automatically arranging freehand sketches of individual categories. 
Ge \etal \cite{ge2022exploring} introduced their own semi-synthetic scene sketch dataset and adopted a DeepLab-v2 \cite{chen2017deeplab} architecture to the scene sketch segmentation task. 
SketchSeger \cite{yang2023scene} proposed an encoder-decoder model based on hierarchical Transformers, trained with a stroke-based cross-entropy loss on semi-synthetic scene sketches formed by combining sketches from the QuickDraw dataset \cite{ha2017neural}.
Zhang \etal \cite{zhang2022stroke} proposed an RNN-GCN-based architecture trained on annotated freehand scene sketches.
However, neither the dataset nor the code have been released.
We do not require stroke-level information or pixel-wise segmentation of the training data, and leverage the FS-COCO dataset \cite{chowdhury2022fs} of freehand sketches with their textual descriptions.

\subsection{ViT-CLIP and Sketch}
% \paragraph{ViT-CLIP and sketch}
We build our encoder on a ViT (Vision Transformer) encoder pre-trained with CLIP (Contrastive Language-Image Pre-training) \cite{radford2021learning}.
CLIP is a model trained on roughly 400 million image-text pairs to embed images and text in a shared space.
It uses ViT as a visual branch (image) encoder.
A ViT encoder pre-trained with CLIP (ViT-CLIP) is used in a range of sketch-related tasks:
sketch and drawing generation \cite{schaldenbrand2021styleclipdraw,frans2022clipdraw,vinker2022clipasso,vinker2022clipascene}, 
2D image retrieval \cite{sangkloy2022sketch,chowdhury2022fs,sain2023clip},
object detection \cite{chowdhury2023can},
3D shape retrieval \cite{schlachter2022zero,le2023sketchanimar,yao2022improving,lee2023learning,berardi2023fine}, 3D shape generation \cite{zheng2023locally}. 

While some works use ViT-CLIP purely pre-trained on images, many fine-tune the encoder on sketches for downstream tasks.
Some works fine-tune all weights of the encoder \cite{sangkloy2022sketch,berardi2023fine}, some fine-tune Layer Normalization layers only \cite{chowdhury2022fs}, and some rely on prompt-learning \cite{jia2022visual,zhou2022conditional} or the combination of the latter two \cite{sain2023clip,chowdhury2023can}.  
In our work, we also rely on fine-tuning with visual prompt learning and Layer Normalization layers updates. 
Unlike previous methods targeting sketch inputs, we additionally leverage a two-path ViT architecture, inspired by Li \etal \cite{li2023clip}.

% \vspace{-0.635cm}

\section{Method}
\label{sec:method}
As we mention in the introduction, we build a sketch encoder such that the semantic meaning of individual stroke pixels can be inferred from its feature embeddings.   
Building on the ViT encoder, pre-trained CLIP \cite{radford2021learning} model, we fine-tune a modified encoder architecture with a network consisting of two levels: Holistic scene understanding and individual category recognition. 
We start by describing the first level of our network (\cref{fig:pipeline} I.) and introducing the architecture of our visual encoder (\cref{fig:pipeline} c.). 
We then describe our strategy to improve the model's ability to understand individual categories (\cref{fig:pipeline} II.).

\subsection{Holistic Scene Sketch Understanding}
The architecture in the first level (\cref{fig:pipeline}: I. Holistic) is similar to the architecture of the CLIP model \cite{li2023clip}. 
We freeze the weights of the textual encoder and fine-tune the modified architecture of the vision encoder (\cref{sec:visual_encoder}). 
The CLIP model is trained with a contrastive loss, ensuring that the embedding of images and corresponding captions are closer in space than embeddings of images and captions of other images. 
While our training has a similar goal, we train with a triplet loss with hard triplet mining, as we found it to be more beneficial with the batch size we use:
\begin{equation}
      \begin {aligned} \mathcal {L}_{N_T\_glbl} = 
      & \frac{1}{N_T} \sum_{i=1}^{N_T} \max\{||\texttt{VST}_i - \texttt{CST}_i^{+}|| \\ 
      & -||\texttt{VST}_i - \texttt{CST}_{j}^{-}||+m, \quad 0\}. 
      \end {aligned} 
      \label{eq:loss_global}
\end{equation}
Here, a holistic visual scene sketch embedding  $\texttt{VST}$ (Visual Scene Token) serves as an anchor. 
An encoding of the matching sketch caption $\texttt{CST}^{+}$ (Caption Scene Token) serves as a positive sample, and an encoding of the most dissimilar scene caption serves as a negative sample $\texttt{CST}^{-}$. 
We set the margin $m$ to a commonly used value of $0.3$. 
The number of triplets $N_T$ is equal to the number of samples in a batch.

\subsubsection{Visual encoder}
% \paragraph{Visual encoder}
\label{sec:visual_encoder}
The input scene sketch is divided into non-overlapping patches, which are flattened and linearly projected into the feature space. Concatenating with positional encodings, we obtain one token $P_k \in \mathbb{R}^{1\times d}$ per patch. 
Additionally, we add a set of learnable tokens, $V_s$, referred to as \emph{visual prompts} \cite{darcet2023vision}.
Finally, these tokens are also augmented with a special token that encodes holistic sketch meaning, $\texttt{VST}$ (Visual Scene Token). Note that in the context of classification, a $\texttt{CLS}$ token has a similar role to our $\texttt{VST}$ token.
Therefore, the input to the vision encoder is $X=[\texttt{VST},P_1,...,P_K,V_1,...,V_S] \in \mathbb{R}^{N_X\times d}$, where $N_X =1+K+S$.

\begin{figure}[t]
  \centering
  % \fbox{\rule{0pt}{2in} \rule{0.9\linewidth}{0pt}}
   \includegraphics[width=\linewidth]{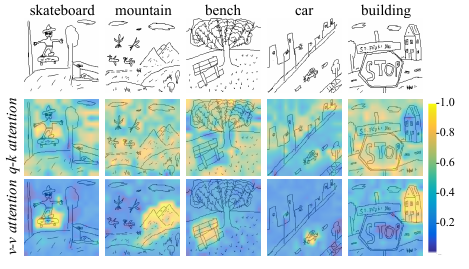}
    \vspace{-0.7cm}
   \caption{Comparison of similarity maps obtained with classical attention computation (\emph{q-k attention}) in the second row, with the ones obtained from \emph{v-v attention}, given by \cref{eq:attnvv}.}
   \label{fig:v-v attention}
   \vspace{-0.2cm}
\end{figure}

% \paragraph{Attention computation in the visual encoder}
\paragraph{Attention computation}
\label{subsec:v-v attention}

It was observed by Li et al.~\cite{li2023clip} that CLIP-predicted similarity maps between image and text features emphasize background regions rather than areas that correspond to a category in the text embedding. 
To address this issue, they proposed to use 
% an alternative configuration of the visual transformer that 
an instance of self-self attention called v-v attention, which 
does not require training or fine-tuning the original model.
Li \etal.~\cite{li2023clip}, and later Bousselham \etal.~\cite{bousselham2023grounding_} demonstrated that this leads to improved performance in open vocabulary segmentation tasks:
Self-self-attention reinforces the similarity of tokens already close to each other (\eg representing the same object), which leads to a clearer separation in the feature space, thereby improving the segmentation quality.

We performed a similar experiment with CLIP features for sketch inputs: The similarity maps in the second row of \cref{fig:v-v attention} show the poor ability of CLIP features to identify target categories.
Therefore, we follow 
% Li et al.~
\cite{li2023clip} and use their two-path configuration of the vision transformer. 
However, we use it not only for inference but also incorporate this two-path configuration directly into our network training, as we find it more beneficial. We provide a detailed analysis in \cref{sec:components}.

The first path represents the original vision encoder where identical blocks are repeated $L$ times. 
Each block consists of \emph{Layer Normalization (LN)}, followed by \emph{Multi-Head Self Attention (MHSA)}, another \emph{LN} and \emph{Fead Forward Network (FFD)}. 

The second path blocks contain a modified attention computation in \emph{MHSA}, dubbed as \emph{v-v self-attention}, where
\emph{Keys} and \emph{Queries} are ignored, and self-attention is computed using only \emph{Values, $V \in \mathbb{R}^{N_X \times d}$}:
\begin{equation}
    \texttt {s-attn}(V,V,V) =  \texttt {softmax}\left(V  V^{T} / \sqrt{d}\right)  V.
    \label{eq:attnvv}
\end{equation}
In addition, blocks in the second path do not include the second \emph{LN} and \emph{FFN} layers. 
Finally, in the second path, the input to the \emph{v-v multi-head attention} is always the features from the original path.
We use the output from the second path during training and inference. 

As shown in \cref{fig:v-v attention} third row, the v-v attention results in feature representations that accurately represent distinct semantic entities present in the scene sketch.

%%%%%%%%%%%%%%%%%%%%%%%%%%%%%%%%%%%%%%%%%
%Categories disentanglement
%%%%%%%%%%%%%%%%%%%%%%%%%%%%%%%%%%%%%%%%%

\subsection{Categories Disentanglement}
\label{sec:disentaglement}
Given the sketch caption we automatically identify individual categories and generate a set of textual prompts of the form \emph{``A sketch of *"} (\cref{fig:pipeline}b.).
Each of these textual category prompts is encoded with the CLIP text encoder into $ \texttt{CCT} \in \mathbb{R}^{1 \times d}$ (Caption Category Token).

We then compute the per-patch cosine similarity $M^c_k$ between the class embeddings $CCT$ and the scene sketch patch embeddings $H_k$, defined as:
\begin{equation}
\label{eq:similarity}
M^c_k = \frac {\texttt{CCT}^c \cdot {H_k^T}}{|\texttt{CCT}^c| |H_k^T|},
\end{equation}
where $k \in [1,K]$ is the patch index and $c \in [1,N_c]$ is an index of a category (\eg\emph{trees}).
The resulting similarity matrix $M^{c} \in \mathbb{R}^{K \times N_c}$ represents the category label probabilities for each individual patch (\cref{fig:pipeline}d.). 
To generate a pixel-level similarity map, 
we reshape each $M^{c}$ and then upscale to the dimensions of the original scene sketch using bi-cubic interpolation \cite{touvron2021training}.
By multiplying these per category maps with the input scene sketch, as shown in \cref{fig:pipeline}e., we obtain disentanglement into individual sketch categories.

% \subsubsection{Thresholding with a Learnable Parameter}
\paragraph{Thresholding with a learnable parameter}
Only pixels with similarity scores above a certain threshold are retained at this step (\cref{fig:pipeline}f.). 
We make the threshold learnable, eliminating the need for manual tuning.
More importantly, the threshold value increases over epochs as the model becomes more confident in its predictions, allowing the model to obtain strong disentanglement performance.

\begin{figure}[ht]
  \centering
   \vspace{-0.2cm}
   \includegraphics[width=\linewidth]{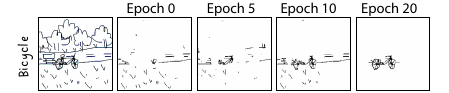}
   \vspace{-0.7cm}
   \caption{Visualization of disentanglement over epochs.}
   \vspace{-0.5cm}
   \label{fig:disentanglement}
\end{figure}

% \subsubsection{Visual Encoder with Cross Attention}
\paragraph{Visual encoder with Cross-Attention}
The features of individual category sketches are extracted with the visual encoder identical to the one used in the holistic scene sketch level understanding of our network, described in \cref{sec:visual_encoder}, up to one difference.  

We enhance the interplay between textual and visual domains through the introduction of cross-attention. 
Namely, in 7th, 10th, and 12th layers in the \emph{MHSA}, we feed $\texttt{CCT}$ token from the textual encoder representing a target category to the linear projection for the queries.
This enables the model to leverage category token embedding from the textual domain to update the sketch token embedding. 
This results in a better text-to-sketch alignment for individual categories and subsequently improves sketch semantic segmentation.
Our ablation study in \cref{tab:ablation_main} underscores the efficacy of this cross-attention strategy.

% \subsubsection{Text-Sketch Category-level Alignment}
\paragraph{Text-sketch category-level alignment}
We train with a triplet loss, $\mathcal {L}_{T\_ctgr}$, so that the category sketch embedding, $\texttt{VCT}$ (Vision Category Token), is used as an anchor, the matching embedding of the category prompt is used as a positive sample and the embedding of the prompt of the most dissimilar category is used as negative. 
We use the $\texttt{VCT}$ from multiple encoder layers: $l_{7},l_{10},l_{12}$.

\subsection{Efficient CLIP fine-tuning}
The two levels (holistic and category) are trained jointly, using the total loss
\begin{equation}
      \mathcal {L} = \mathcal {L}_{T\_glbl} + \mathcal {L}_{T\_ctgr}.
\end{equation}

We leverage the generalization properties of the pre-trained foundation model through careful fine-tuning.
We freeze all the weights apart from weights of \emph{LN}, as was proposed in \cite{frankle2020training}, 
and we use learnable visual prompts, as was proposed in \cite{jia2022visual}. We introduced visual prompts in \cref{sec:visual_encoder}. 
We also train linear layers which take part in cross-attention computation.

\subsection{Inference}
\label{sec:inference}
Our network design allows segmentation for different sets of categories.
Given a desirable set of categories for a given sketch, we obtain sketch segmentation by applying all the steps of our network up to the calculation of pixel-category similarities (\cref{fig:pipeline}e.), followed by upscaling
of similarity maps for each category, as discussed in \cref{sec:disentaglement}.
To assign segmentation results we assign to each pixel a label that yields the highest similarity value across category similarity maps $M^c_i$, where $i$ is an index of a category. 

If we want to isolate just a few categories in the sketch, we can use the thresholding strategy that we use during training to isolate the pixels of a given category (\cref{fig:pipeline}f.). 
We used this strategy to obtain visualizations in \cref{fig:teaser}, with a threshold value of $0.71$ that we found to be optimal on the test set of sketches.
We do not use the learned value from the training, as during training the model does not have to select all the pixels of the given category, but only those that are sufficient to confidently predict the category label.
We provide an in-depth discussion in the supplemental.

\begin{figure*}
  \centering
  \vspace{-0.3cm}
    \includegraphics[width=\linewidth]{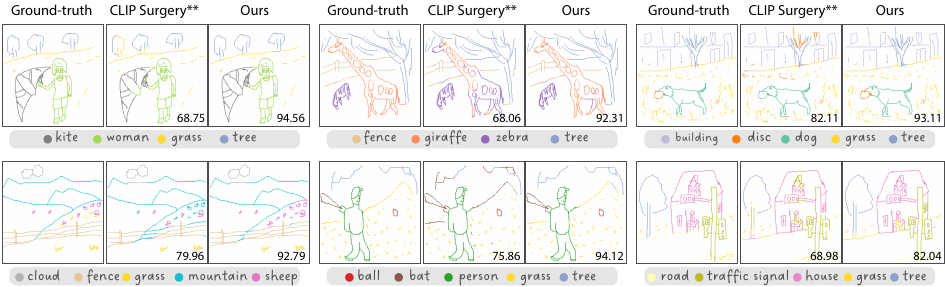}
    \vspace{-0.7cm}
   \caption{Visual comparison of our method with \emph{CLIP Surgery\textsuperscript{$\star$$\star$}}. \emph{CLIP Surgery\textsuperscript{$\star$$\star$}} represents the fine-tuned ViT from the CLIP model with v-v self-attention introduced at both training and inference stages. 
   The numbers show \emph{Acc@P} values.}
   \label{fig:r1}
\end{figure*}
\newcommand{\myparagraph}[1]{\vspace{3pt}\noindent{\bf #1}}

\section{Experiments}
\subsection{Training and Test Data}
For training and testing, we use the sketch-caption pairs from the FS-COCO \cite{chowdhury2022fs} dataset. 
The dataset comprises 10,000 sketch-caption pairs, associated with reference images from the MS-COCO \cite{lin2014microsoft} dataset.  
The sketches are drawn from memory by 100 non-expert participants. 
The reference image was shown for 60 seconds, followed by a 3-minute sketching window. 

\myparagraph{Training/Validation/Test splits} We first selected 500 sketches with distinct styles from five participants.
We then randomly sample 5 sketches from each of the remaining 95 participants for validation (a total of 475 sketches).
We use the remaining 9025 sketches for training. 

\myparagraph{Annotations} One of the co-authors manually annotated test and validation sketches, relying on reference images and category labels from the MS-COCO \cite{lin2014microsoft}  dataset. 
We assign each stroke a unique category label. 
Candidate category labels are extracted from MS-COCO image captions rather than sketch captions to obtain richer \emph{`ground-truth'} annotations. 
Our test set contains 185 different object classes, with an average of 3.54 objects per sketch.

\subsection{Evaluation Metrics}
We use standard metrics, commonly used in sketch segmentation literature \cite{huang2014data,wu2018sketchsegnet,zhang2022stroke}.\\
\textbf{Mean Intersection over Union ($mIoU$):} 
evaluates the average of the ratios between the intersection and the union of ground truth and predicted labels over all categories.\\
\textbf{Pixel Accuracy ($Acc@{P}$):} measures the ratio of correctly labeled pixels to the total pixel count in a sketch.\\
\textbf{Stroke Accuracy ($Acc@{S}$):} evaluates the percentage of correctly classified strokes to total strokes per sketch. A stroke label is determined by its most frequent pixel label.

\subsection{Implementation Details}
\label{sec:implementation_details}
We implemented our method in PyTorch and trained on two $24$GB Nvidia RTX A$5000$ GPUs. 
We built on CLIP \cite{radford2021learning} with a ViT backbone using ViT-B/16 weights.  
The input sketch image size is set as $224 \times 224$. 
We use 3 learnable visual prompts.
We use AdamW optimizer with a learning rate of $10^{-6}$, and train the model for $20$ epochs with a batch size of $16$.  
We pick a checkpoint based on the \emph{mIoU} performance on the validation set. 
We provide more discussion on the checkpoint choice in the supplemental.

\subsection{Comparison against state-of-the-art}
% \subsubsection{Comparison with methods relying on the availability of pixel-level annotations}
\subsubsection{Comparison with fully-supervised methods}
We first compare with several recent methods for image segmentation that similarly to us utilize either CLIP as a backbone: \emph{DenseCLIP} \cite{rao2022denseclip} and \emph{ZegCLIP} \cite{zhou2023zegclip}, or more recent foundational backbones Grounding-DINO \cite{liu2023grounding} and SAM \cite{kirillov2023segany}, used in \emph{Grounded-SAM} \cite{groundedSAM}. These methods require pixel-level annotated examples, and therefore can not be fine-tuned on our training data.  
We also compare to a recent fully supervised method \emph{LDP} (Local Detail Perception) \cite{ge2022exploring} for scene sketch semantic segmentation, which is trained on a dataset of semi-synthetic sketches. 
Such sketches are obtained as a superposition of freehand category-level sketches. 
% We refer to this method as \emph{LDP} (Local Detail Perception).
%
% \cref{tab:supervised} shows that neither of the considered methods generalizes well to freehand scene sketches.
\cref{tab:supervised} shows that neither of the these methods generalizes well to freehand scene sketches.
% with a large number of categories. 

\begin{table}[ht]
  \centering
  % \resizebox{\columnwidth}{!}{  
  % \small{
  \begin{tabular}{@{}lcccc@{}}
    \toprule
     Methods & $mIoU$ & $Acc@{P}$ & $Acc@{S}$   \\
    \midrule
    ZegCLIP \cite{zhou2023zegclip}&  $15.45$ & $32.48$ & $35.21$ \\ 
    DenseCLIP \cite{zhou2021denseclip}&$28.22$ & $50.62$ & $50.25$\\
    Grounded-SAM \cite{groundedSAM}&  $32.21$ & $50.12$ & $50.02$ \\
    \midrule
    LDP \cite{ge2022exploring} &  $33.04$ & $56.23$ & $56.71$ \\
    \midrule
    \textbf{Ours} &  $\textbf{73.48}$ & $\textbf{85.54}$ & $\textbf{87.02}$ \\
    \bottomrule
  \end{tabular}
  % }
  %  }
  \vspace{-0.2cm}
  \caption{Comparison of our method against state-of-the-art fully supervised sketch method and image segmentation methods, relying on the availability of pixel-level annotations, on our test set of freehand sketches from the FS-COCO dataset.}
    \label{tab:supervised}
    \vspace{-0.4cm}
\end{table}

\newcommand{\myparagraphSeven}[1]{\vspace{8pt}\noindent{\bf #1}}
\subsubsection{Comparison with language-supervised methods}
Next, we compare with several recent methods targeting semantic segmentation with ViT encoders and image-text supervision: \emph{GroupViT} \cite{xu2022groupvit} and \emph{SegCLIP} \cite{luo2022segclip}. 
Additionally, we compare with CLIP \cite{radford2021learning}, as well as CLIP Surgery \cite{li2023clip} that introduced the usage of \emph{v-v-attention} at inference time.

\myparagraphSeven{Zero-shot} 
In \cref{tab:results_main}, we first compare the performance of our method with the zero-shot performance of these methods. 
It shows that image segmentation methods do not generalize well to freehand sketches. 

\myparagraphSeven{Fine-tuning} 
We fine-tune each of the methods on our training set, by updating all their weights.  
Since such fine-tuning might be sensitive to a learning-rate choice, we perform several runs with several settings of learning rate parameters. 
We chose the setting for each method that results in the best performance on our validation set. 
The fine-tuned methods are marked with stars.

\cref{tab:results_main} shows that our method outperforms all considered baselines, and surpasses the best-performing baseline \emph{CLIP Surgery\textsuperscript{$\star$$\star$}} by a substantial margin of $13.5$, $9.9$ and $5.9$ points in $mIoU$ score, $Acc@{P}$ and $Acc@{S}$, respectively. 
In \cref{sec:components}, we evaluate various elements of our architecture and their contribution to overall performance.

\begin{table}[ht]
\centering
\resizebox{\columnwidth}{!}{%
\begin{tabular}{lllcc}
\toprule
 & Methods & $mIoU$ & $Acc@{P}$ & $Acc@{S}$ \\ \hline
 \parbox[t]{2mm}{\multirow{4}{*}{\rotatebox[origin=c]{90}{Zero-shot}}} & CLIP \cite{radford2021learning} & $17.33$ & $28.82$ & $27.15$ \\
 & GroupViT \cite{xu2022groupvit} & $38.25$ & $61.39$ & $60.07$ \\
 & SegCLIP \cite{luo2022segclip} & $38.14$ & $61.45$ & $65.56$ \\
 & CLIP\_Surgery \cite{li2023clip} & $52.63$ & $72.47$ & $75.17$ \\ \hline
\parbox[t]{2mm}{\multirow{5}{*}{\rotatebox[origin=c]{90}{Fine-tuned}}}& CLIP\textsuperscript{$\star$} & $22.86$ & $33.41$ & $32.64$ \\
 & GroupViT\textsuperscript{$\star$} & $45.71$ & $66.21$ & $66.89$ \\
 & SegCLIP\textsuperscript{$\star$} & $49.26$ & $69.87$ & $73.64$ \\
 & CLIP\_Surgery\textsuperscript{$\star$} & $48.74$ & $65.38$ & $68.78$ \\
 & CLIP\_Surgery\textsuperscript{$\star$$\star$} & $59.98$ & $78.68$ & $81.11$ \\ \hline
 & \textbf{Ours} & $\textbf{73.48}$ & $\textbf{85.54}$ & $\textbf{87.02}$ \\  \bottomrule
\end{tabular}%
}
  \vspace{-0.2cm}
  \caption{Comparison of our method against state-of-the-art language supervised image segmentation methods on our test set of sketches from the FS-COCO dataset.
  The fine-tuned methods on our training set of freehand sketches are marked with stars.
  \emph{CLIP Surgery\textsuperscript{$\star$}} represents the fine-tuned CLIP model with v-v self-attention introduced only at inference stages.
\emph{CLIP Surgery\textsuperscript{$\star$$\star$}} represents the fine-tuned model with v-v self-attention introduced at both training and inference stages.}
  \label{tab:results_main}
\end{table}

\cref{fig:r1} shows the qualitative comparison between our method and the \emph{CLIP Surgery\textsuperscript{$\star$$\star$}}. 
We provide additional visual comparisons in the supplemental.

\subsubsection{Generalization ability of our method}
\label{sec:generalization}
Next, we evaluate our method on an additional dataset of 50 freehand sketches provided and annotated by Ge \etal~\cite{ge2022exploring}.
\cref{tab:zs_freehand} shows that our model again demonstrates superior performance on this dataset over the method \cite{ge2022exploring}, fully supervised on semi-synthetic sketches.
We do not compute \emph{Acc@S} as sketches are only available as bitmap images. 
This experiment highlights that short language captions can be efficiently used for training, eliminating the need for expensive and time-consuming per-pixel annotations. 
\begin{table}[ht]
  \centering
    \begin{tabular}{lrr}
        \toprule
        Method & \multicolumn{1}{c}{$mIoU$} & \multicolumn{1}{c}{$Acc@P$}  \\ 
         \midrule
        LDP \cite{ge2022exploring} & $37.16 $ & $78.84$ \\
        Ours & $\textbf{53.94}$ & $\textbf{81.63}$ \\ 
        \bottomrule
    \end{tabular}%
    \vspace{-0.3cm}
  \caption{Comparison on the freehand sketches from \cite{ge2022exploring}.}
  \label{tab:zs_freehand}
\end{table}

The lower \emph{mIoU} values on these sketches than on FS-COCO sketches can be explained by (1) on larger average number of categories in them  ($5.74$ categories per sketch) than in our FS-COCO test set ($3.54$ categories per sketch); (2) domain gap.
The sketches from \cite{ge2022exploring}
contain symbolic representations of objects (see the inset
\setlength{\columnsep}{8pt}
\begin{wrapfigure}[5]{l}{2cm}
\centering
\vspace{-0.4cm}
\includegraphics[width=2cm]{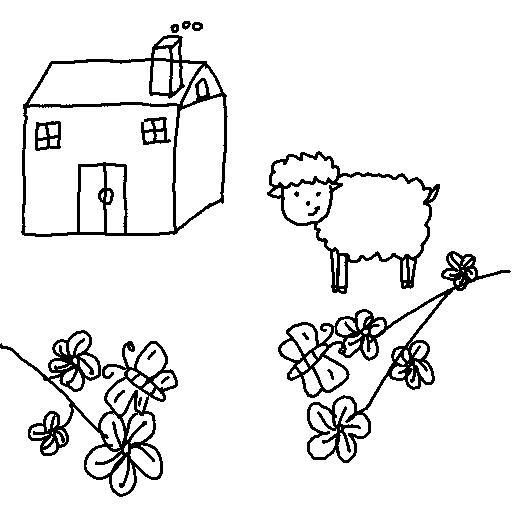}
% \hspace{-1.2cm}
% \hspace{-0.4cm}
% \vspace{}
   % \caption{Sample freehand sketch from \cite{ge2022exploring}.}
% \label{fig:dataset_size}
% \vspace{-10pt}
\end{wrapfigure}
on the left) and look more like a superposition of sketches that can be found in the \emph{QuickDraw} \cite{ha2017neural} dataset rather than holistic scene sketches. 
We analyze challenging scenarios for our method in \cref{sec:challenges}.

\subsection{Ablation Study}
\subsubsection{Importance of individual components}
\label{sec:components}
We perform an ablation analysis to assess the importance of each component in our architecture. 
\cref{tab:ablation_main} shows the performance of the complete model with individual elements removed. 
We discuss them in order of impact on overall performance. 

\myparagraphZ{v-v attention} First, we show the importance of the v-v attention, by substituting our dual path v-v attention-based ViT encoder with the original configuration used in the CLIP model (\textbf{w/o v-v attention}).

\myparagraphZ{Two-level network architecture} 
We keep only the first level of holistic scene understanding of the network (\cref{fig:pipeline} I.).
This architecture is similar to \emph{CLIP Surgery\textsuperscript{$\star$$\star$}}, but is supervised with the triplet loss and is fine-tuned using \emph{learnable visual prompts} and updates only \emph{LN} layers.
\cref{tab:ablation_main} \emph{(w/o category-level)} confirms that two-level network architecture, along \emph{v-v attention}, is central to the superiority of our model. 

\myparagraphZ{Thresholding}
We perform an experiment where instead of thresholding we weight each pixel according to cosine similarity scores in $M^c$ maps (\cref{tab:ablation_main} \emph{(w/o thresholding)}). 
The learnable threshold more efficiently filters out irrelevant pixels, forcing the model to learn superior disentanglement of individual categories.

\myparagraphZ{Holistic scene encoding}
Removing the global loss, given by \cref{eq:loss_global}, similarly results in the performance drop \emph{(w/o Global Loss)}. 
This shows the mutual importance of the two levels of our network. 

\myparagraphZ{Cross-Attention} Cross attention also substantially contributes to performance. 
If we use a ViT encoder at the second level of the network (category level), identical to the one used at the first level (holistic level) (\cref{fig:pipeline}c.), then the performance drops by a noticeable $3.35$ points in the \emph{mIoU} score (\cref{tab:ablation_main} \emph{(w/o cross-attention)}).

\myparagraphZ{Multi-layer features in the triplet loss}
\cref{tab:ablation_main} \emph{(w/o cross-attention)} shows that using features from multiple layers ({$l_{7}, l_{10}, l_{12}$}) in the category-level triplet loss is beneficial over using only the features from the last layer ($l_{12}$).

\begin{table}[th]
  \centering
  \small{
  \begin{tabular}{@{}lccc@{}}
    \toprule
    Model & $mIoU$ & $Acc@{P}$ & $Acc@{S}$   \\
    \midrule
    w/o v-v attention & $43.55$ & $58.09$ & $59.03$ \\ 
    w/o category-level & $65.03$ & $79.35$ & $81.82$ \\
    w/o thresholding & $66.93$ & $81.06$ & $82.56$ \\
    w/o global Loss & $ 69.06$ & $81.35$ & $83.68$ \\
    w/o cross-attention & $70.13$ & $82.86$ & $85.26$ \\
    w/o multi-layer Loss & $71.29$ & $83.04$ & $86.13$\\
    \midrule
    \textbf{Ours-full} & $\textbf{73.48}$ & $\textbf{85.54}$ & $\textbf{87.02}$ \\
    \bottomrule
    \vspace{-0.4cm}
  \end{tabular}
  }
  \vspace{-0.2cm}
  \caption{Ablation of the role of individual components of our model. See \cref{sec:components} for details.}
  \label{tab:ablation_main}
  \vspace{-0.2cm}
\end{table}

\subsubsection{Efficient fine-tuning}
\begin{figure}[ht]
  \centering
   % \vspace{-0.2cm}
   \includegraphics[width=\linewidth]{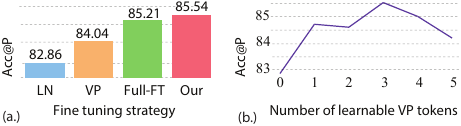}
    \vspace{-0.4cm}
   \caption{Evaluation of alternative fine-tuning strategies (a.) and the impact of the number of learnable tokens on segmentation accuracy (b.). \emph{LN} means that only \emph{LN} layers are fine-tuned; \emph{VP} means that only learnable Visual Prompt tokens are used; \emph{Full-FT} means that all weights of ViT are fine-tuned.}
   % \vspace{-0.5cm}
   \label{fig:ft}
\end{figure}

\cref{fig:ft} shows the comparison of different fine-tuning strategies. 
We obtain the best results by combining fine-tuning of \emph{LN} (Layer Normalization) layers and the addition of $3$ learnable tokens. 
Adding more or less tokens degrades the performance \cref{fig:ft}b.

\section{Human-Model Alignment}
\label{sec:human_model_align}
\begin{figure}[t]
  \centering
   \includegraphics[width=\linewidth]{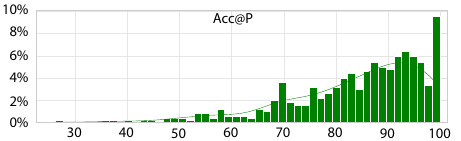}
   \vspace{-0.6cm}
   \caption{Histogram of Acc@P values for our method on 500 sketches from our FS-COCO test set.}
   \label{fig:hist_acc}
\end{figure}

\cref{fig:hist_acc} shows that for the majority of sketches in our test set from the FS-COCO dataset, our model correctly labels more than $80\%$ pixels. 

In this section, we investigate (1) which sketches are likely to get low segmentation accuracy and (2) how the prediction of our model compares with human observers across different groups of sketches.

\subsection{Sketch Groups}
\label{sec:challenges}
We identified four distinct sketch groups that are challenging for our model: 
(1) \textbf{Ambiguous sketches}: sketches where it might be hard even for a human observer to understand an input sketch; 
(2) \textbf{Interchangeable categories}: sketches containing multiple objects with labels that can interchange each other, like \emph{`tower'} and \emph{`building'}, or \emph{`girl'} and \emph{`man'};
(3) \textbf{Correlated categories}: sketches with categories that typically co-occur in scenes, \eg, \emph{`train'-`railway'} and \emph{`airplane'-`runway'}; 
and 
(4) \textbf{Numerous-categories}: sketches with six or more categories. 

We supplement these four groups with sketches where our model labels correctly more than $80\%$ of pixels: (5) \textbf{Strong performance}.

\subsection{User Study Setting}
\paragraph{Data}
We sample 5 sketches for each of the first 4 categories and 10 sketches for the 5th category.  
We visualize selected sketches in the supplemental material.

\myparagraphZ{Participants}
We recruited 25 participants ($14$ male). 
Each participant was randomly assigned 6 sketches: 1 from each of the first 4 groups and 2 from the 5th group, such that every sketch was annotated by five unique participants.

\myparagraphZ{Study Procedure}
Participants were presented with one sketch and one object category at a time and were not able to see their previous annotations. 
Sketch-category pairs were interlaced, to reduce the effect of memorizing their previous annotation on a certain sketch. 
The annotation interface enabled precise pixel-level segmentation by allowing participants to ``paint'' over each sketch using a brush with an adjustable radius. 
Participants could also use the eraser to correct erroneous annotations.
Once a participant has moved to a new sketch-category pair, they were not able to change their previous annotations.

\subsection{User Study Analysis and Future Work}
\paragraph{\emph{`Human'} segmentation} For each sketch, we generate one \emph{`human'} segmentation using a majority vote. 
For each pixel and each label, we computed the percentage of annotators that assigned a given label. 
We then assigned to each pixel the label that was provided most frequently to that pixel by different annotators.
In cases where there were multiple labels were provided equally often for a pixel, we randomly sampled one of these labels.

\begin{figure}[t]
\vspace{-0.0cm}
  \centering
   \includegraphics[width=\linewidth]{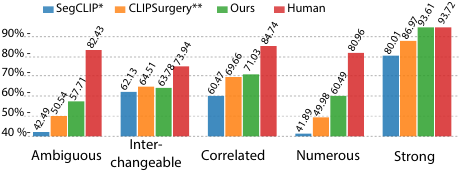}
   \vspace{-0.7cm}
   \caption{Comparison of the percentage of correctly predicted pixels (\emph{Acc@P}) by different models and human observers across five distinct sketch categories, introduced in \cref{sec:challenges}. 
   \vspace{-0.0cm}
   }
   \label{fig:human_vs_model}
\end{figure}

\vspace{0.6cm}
\myparagraphZ{Analysis}
First, we observed that on sketches that did not fall into any of the challenging categories, our model almost reaches human-level performance, with a negligible gap of $0.11$ points on average (\cref{fig:human_vs_model} Strong).  

% Interestingly, given a label humans can correctly identify sketch pixels even in the presence of ambiguity (\cref{fig:human_vs_model} Ambiguous). 
\cref{fig:human_vs_model} Ambiguous shows that, given a label, humans can correctly identify sketch pixels even in the presence of ambiguity. 
While none of the models currently match human performance on \emph{ambiguous sketches}, our model surpasses the other methods by a noticeable margin, demonstrating the effectiveness of our two-level training architecture. 

The performance across \emph{semantically interchangeable categories} is uniform amongst the three language-supervised models.
This potentiality can be alleviated by proposing solutions that assign labels jointly.   

On sketches with \emph{correlated categories} our model and ClipSurgery\textsuperscript{$\star$$\star$} perform similarly, highlighting the inherent limitation of training using language supervision.
For a few such categories, one might need to further fine-tune the model relying on sketches of isolated categories. 

Our model represents a substantial improvement over current alternatives, surpassing them by more than $10$ points.  
Future work should seek to improve alignment with human sketch understanding, especially on sketches with more than six categories (\cref{fig:human_vs_model} Numerous). 

\section{Conclusion}
\begin{spacing}{1.03}
While focusing on the task of sketch segmentation, we introduced a strategy to train a ViT encoder that results in the feature space with good semantic disentanglement.
Such feature spaces contribute towards improving machine understanding of abstract freehand sketches and underpin a range of downstream tasks such as communication and creative pipelines. 
In light of the latter, it can enable more potent tools for conditional generation and retrieval. 
In psychology, sketches are used to analyze cognitive functions.
This can be facilitated by the availability of robust sketch understanding tools.  
Importantly, we for the first time demonstrated how language supervision can be used for the task of scene sketch segmentation. 
Finally, we conducted a comprehensive analysis of our model's performance, identifying research directions to further align the understanding of sketches by humans and machines.
% The code and all annotations will be made available upon acceptance.
\end{spacing}
{
    \small
    \bibliographystyle{ieeenat_fullname}
    \bibliography{main}

\begin{thebibliography}{76}
\providecommand{\natexlab}[1]{#1}
\providecommand{\url}[1]{\texttt{#1}}
\expandafter\ifx\csname urlstyle\endcsname\relax
  \providecommand{\doi}[1]{doi: #1}\else
  \providecommand{\doi}{doi: \begingroup \urlstyle{rm}\Url}\fi

\bibitem[Badrinarayanan et~al.(2017)Badrinarayanan, Kendall, and Cipolla]{badrinarayanan2017segnet}
Vijay Badrinarayanan, Alex Kendall, and Roberto Cipolla.
\newblock Segnet: A deep convolutional encoder-decoder architecture for image segmentation.
\newblock \emph{IEEE transactions on pattern analysis and machine intelligence}, 39\penalty0 (12):\penalty0 2481--2495, 2017.

\bibitem[Berardi and Gryaditskaya(2023)]{berardi2023fine}
Gianluca Berardi and Yulia Gryaditskaya.
\newblock Fine-tuned but zero-shot 3d shape sketch view similarity and retrieval.
\newblock In \emph{Proceedings of the IEEE/CVF International Conference on Computer Vision}, pages 1775--1785, 2023.

\bibitem[Bousselham et~al.(2023)Bousselham, Petersen, Ferrari, and Kuehne]{bousselham2023grounding_}
Walid Bousselham, Felix Petersen, Vittorio Ferrari, and Hilde Kuehne.
\newblock Grounding everything: Emerging localization properties in vision-language transformers.
\newblock \emph{arXiv preprint arXiv:2312.00878}, 2023.

\bibitem[Caesar et~al.(2018)Caesar, Uijlings, and Ferrari]{caesar2018coco}
Holger Caesar, Jasper Uijlings, and Vittorio Ferrari.
\newblock Coco-stuff: Thing and stuff classes in context.
\newblock In \emph{Proceedings of the IEEE conference on computer vision and pattern recognition}, pages 1209--1218, 2018.

\bibitem[Chan et~al.(2022)Chan, Durand, and Isola]{chan2022drawings}
Caroline Chan, Fredo Durand, and Phillip Isola.
\newblock Learning to generate line drawings that convey geometry and semantics.
\newblock 2022.

\bibitem[Chen et~al.(2023)Chen, Zhu, Qian, Ghanem, Yan, Zhu, Xiao, Elhoseiny, and Culatana]{chen2023exploring}
Jun Chen, Deyao Zhu, Guocheng Qian, Bernard Ghanem, Zhicheng Yan, Chenchen Zhu, Fanyi Xiao, Mohamed Elhoseiny, and Sean~Chang Culatana.
\newblock Exploring open-vocabulary semantic segmentation without human labels.
\newblock \emph{arXiv preprint arXiv:2306.00450}, 2023.

\bibitem[Chen et~al.(2017{\natexlab{a}})Chen, Papandreou, Kokkinos, Murphy, and Yuille]{chen2017deeplab}
Liang-Chieh Chen, George Papandreou, Iasonas Kokkinos, Kevin Murphy, and Alan~L Yuille.
\newblock Deeplab: Semantic image segmentation with deep convolutional nets, atrous convolution, and fully connected crfs.
\newblock \emph{IEEE transactions on pattern analysis and machine intelligence}, 40\penalty0 (4):\penalty0 834--848, 2017{\natexlab{a}}.

\bibitem[Chen et~al.(2017{\natexlab{b}})Chen, Papandreou, Schroff, and Adam]{chen2017rethinking}
Liang-Chieh Chen, George Papandreou, Florian Schroff, and Hartwig Adam.
\newblock Rethinking atrous convolution for semantic image segmentation.
\newblock \emph{arXiv preprint arXiv:1706.05587}, 2017{\natexlab{b}}.

\bibitem[Cho et~al.(2021)Cho, Mall, Bala, and Hariharan]{cho2021picie}
Jang~Hyun Cho, Utkarsh Mall, Kavita Bala, and Bharath Hariharan.
\newblock Picie: Unsupervised semantic segmentation using invariance and equivariance in clustering.
\newblock In \emph{Proceedings of the IEEE/CVF Conference on Computer Vision and Pattern Recognition}, pages 16794--16804, 2021.

\bibitem[Chowdhury et~al.(2022)Chowdhury, Sain, Bhunia, Xiang, Gryaditskaya, and Song]{chowdhury2022fs}
Pinaki~Nath Chowdhury, Aneeshan Sain, Ayan~Kumar Bhunia, Tao Xiang, Yulia Gryaditskaya, and Yi-Zhe Song.
\newblock Fs-coco: towards understanding of freehand sketches of common objects in context.
\newblock In \emph{Computer Vision--ECCV 2022: 17th European Conference, Tel Aviv, Israel, October 23--27, 2022, Proceedings, Part VIII}. Springer, 2022.

\bibitem[Chowdhury et~al.(2023)Chowdhury, Bhunia, Sain, Koley, Xiang, and Song]{chowdhury2023can}
Pinaki~Nath Chowdhury, Ayan~Kumar Bhunia, Aneeshan Sain, Subhadeep Koley, Tao Xiang, and Yi-Zhe Song.
\newblock What can human sketches do for object detection?
\newblock In \emph{CVPR}, 2023.

\bibitem[Darcet et~al.(2023)Darcet, Oquab, Mairal, and Bojanowski]{darcet2023vision}
Timoth{\'e}e Darcet, Maxime Oquab, Julien Mairal, and Piotr Bojanowski.
\newblock Vision transformers need registers.
\newblock \emph{arXiv preprint arXiv:2309.16588}, 2023.

\bibitem[Defferrard et~al.(2016)Defferrard, Bresson, and Vandergheynst]{defferrard2016convolutional}
Micha{\"e}l Defferrard, Xavier Bresson, and Pierre Vandergheynst.
\newblock Convolutional neural networks on graphs with fast localized spectral filtering.
\newblock \emph{Advances in neural information processing systems}, 29, 2016.

\bibitem[Ding et~al.(2022)Ding, Xue, Xia, and Dai]{ding2022decoupling}
Jian Ding, Nan Xue, Gui-Song Xia, and Dengxin Dai.
\newblock Decoupling zero-shot semantic segmentation.
\newblock In \emph{Proceedings of the IEEE/CVF Conference on Computer Vision and Pattern Recognition}, pages 11583--11592, 2022.

\bibitem[Dong et~al.(2023)Dong, Bao, Zheng, Zhang, Chen, Yang, Zeng, Zhang, Yuan, Chen, et~al.]{dong2023maskclip}
Xiaoyi Dong, Jianmin Bao, Yinglin Zheng, Ting Zhang, Dongdong Chen, Hao Yang, Ming Zeng, Weiming Zhang, Lu Yuan, Dong Chen, et~al.
\newblock Maskclip: Masked self-distillation advances contrastive language-image pretraining.
\newblock In \emph{Proceedings of the IEEE/CVF Conference on Computer Vision and Pattern Recognition}, pages 10995--11005, 2023.

\bibitem[Eitz et~al.(2012)Eitz, Hays, and Alexa]{eitz2012humans}
Mathias Eitz, James Hays, and Marc Alexa.
\newblock How do humans sketch objects?
\newblock \emph{ACM Transactions on graphics (TOG)}, 31\penalty0 (4):\penalty0 1--10, 2012.

\bibitem[Frankle et~al.(2020)Frankle, Schwab, and Morcos]{frankle2020training}
Jonathan Frankle, David~J Schwab, and Ari~S Morcos.
\newblock Training batchnorm and only batchnorm: On the expressive power of random features in cnns.
\newblock \emph{arXiv preprint arXiv:2003.00152}, 2020.

\bibitem[Frans et~al.(2022)Frans, Soros, and Witkowski]{frans2022clipdraw}
Kevin Frans, Lisa Soros, and Olaf Witkowski.
\newblock Clipdraw: Exploring text-to-drawing synthesis through language-image encoders.
\newblock \emph{Advances in Neural Information Processing Systems}, 35:\penalty0 5207--5218, 2022.

\bibitem[Fu et~al.(2019)Fu, Liu, Tian, Li, Bao, Fang, and Lu]{fu2019dual}
Jun Fu, Jing Liu, Haijie Tian, Yong Li, Yongjun Bao, Zhiwei Fang, and Hanqing Lu.
\newblock Dual attention network for scene segmentation.
\newblock In \emph{Proceedings of the IEEE/CVF conference on computer vision and pattern recognition}, pages 3146--3154, 2019.

\bibitem[Gao et~al.(2020)Gao, Liu, Xu, Wang, Liu, and Zou]{gao2020sketchycoco}
Chengying Gao, Qi Liu, Qi Xu, Limin Wang, Jianzhuang Liu, and Changqing Zou.
\newblock Sketchycoco: Image generation from freehand scene sketches.
\newblock In \emph{Proceedings of the IEEE/CVF conference on computer vision and pattern recognition}, pages 5174--5183, 2020.

\bibitem[Ge et~al.(2022)Ge, Sun, Song, Ma, and Liao]{ge2022exploring}
Ce Ge, Haifeng Sun, Yi-Zhe Song, Zhanyu Ma, and Jianxin Liao.
\newblock Exploring local detail perception for scene sketch semantic segmentation.
\newblock \emph{IEEE Transactions on Image Processing}, 31, 2022.

\bibitem[GroundedSAM(2023)]{groundedSAM}
GroundedSAM.
\newblock {Grounded-Segment-Anything}.
\newblock https://github.com/IDEA-Research/Grounded-Segment-Anything, 2023.

\bibitem[Ha and Eck(2017)]{ha2017neural}
David Ha and Douglas Eck.
\newblock A neural representation of sketch drawings.
\newblock \emph{arXiv preprint arXiv:1704.03477}, 2017.

\bibitem[H{\"a}hnlein et~al.(2019)H{\"a}hnlein, Gryaditskaya, and Bousseau]{hahnlein2019bitmap}
F H{\"a}hnlein, Y Gryaditskaya, and A Bousseau.
\newblock Bitmap or vector? a study on sketch representations for deep stroke segmentation.
\newblock In \emph{Journ{\'e}es Francaises d'Informatique Graphique et de R{\'e}alit{\'e} virtuelle}, 2019.

\bibitem[Hamilton et~al.(2022)Hamilton, Zhang, Hariharan, Snavely, and Freeman]{hamilton2022unsupervised}
Mark Hamilton, Zhoutong Zhang, Bharath Hariharan, Noah Snavely, and William~T Freeman.
\newblock Unsupervised semantic segmentation by distilling feature correspondences.
\newblock \emph{arXiv preprint arXiv:2203.08414}, 2022.

\bibitem[He et~al.(2023)He, Jamonnak, Gou, and Ren]{he2023clips4}
Wenbin He, Suphanut Jamonnak, Liang Gou, and Liu Ren.
\newblock Clip-s$^4$: Language-guided self-supervised semantic segmentation, 2023.

\bibitem[Huang et~al.(2014)Huang, Fu, and Lau]{huang2014data}
Zhe Huang, Hongbo Fu, and Rynson~WH Lau.
\newblock Data-driven segmentation and labeling of freehand sketches.
\newblock \emph{ACM Transactions on Graphics (TOG)}, 33\penalty0 (6):\penalty0 1--10, 2014.

\bibitem[Hwang et~al.(2019)Hwang, Yu, Shi, Collins, Yang, Zhang, and Chen]{hwang2019segsort}
Jyh-Jing Hwang, Stella~X Yu, Jianbo Shi, Maxwell~D Collins, Tien-Ju Yang, Xiao Zhang, and Liang-Chieh Chen.
\newblock Segsort: Segmentation by discriminative sorting of segments.
\newblock In \emph{Proceedings of the IEEE/CVF International Conference on Computer Vision}, pages 7334--7344, 2019.

\bibitem[Jia et~al.(2022)Jia, Tang, Chen, Cardie, Belongie, Hariharan, and Lim]{jia2022visual}
Menglin Jia, Luming Tang, Bor-Chun Chen, Claire Cardie, Serge Belongie, Bharath Hariharan, and Ser-Nam Lim.
\newblock Visual prompt tuning.
\newblock In \emph{Computer Vision--ECCV 2022: 17th European Conference, Tel Aviv, Israel, October 23--27, 2022, Proceedings, Part XXXIII}, pages 709--727. Springer, 2022.

\bibitem[Kaiyrbekov and Sezgin(2019)]{kaiyrbekov2019deep}
Kurmanbek Kaiyrbekov and Metin Sezgin.
\newblock Deep stroke-based sketched symbol reconstruction and segmentation.
\newblock \emph{IEEE computer graphics and applications}, 40\penalty0 (1):\penalty0 112--126, 2019.

\bibitem[Kirillov et~al.(2023)Kirillov, Mintun, Ravi, Mao, Rolland, Gustafson, Xiao, Whitehead, Berg, Lo, Doll{\'a}r, and Girshick]{kirillov2023segany}
Alexander Kirillov, Eric Mintun, Nikhila Ravi, Hanzi Mao, Chloe Rolland, Laura Gustafson, Tete Xiao, Spencer Whitehead, Alexander~C. Berg, Wan-Yen Lo, Piotr Doll{\'a}r, and Ross Girshick.
\newblock Segment anything.
\newblock \emph{arXiv:2304.02643}, 2023.

\bibitem[Le et~al.(2023)Le, Nguyen, Le, Nguyen, Huynh, Do, Le, Tran, Hoang-Xuan, Nguyen-Ho, et~al.]{le2023sketchanimar}
Trung-Nghia Le, Tam~V Nguyen, Minh-Quan Le, Trong-Thuan Nguyen, Viet-Tham Huynh, Trong-Le Do, Khanh-Duy Le, Mai-Khiem Tran, Nhat Hoang-Xuan, Thang-Long Nguyen-Ho, et~al.
\newblock Sketchanimar: Sketch-based 3d animal fine-grained retrieval.
\newblock \emph{arXiv preprint arXiv:2304.05731}, 2023.

\bibitem[Lee et~al.(2023)Lee, Hwang, Go, Choi, Kim, and Zhang]{lee2023learning}
Hyundo Lee, Inwoo Hwang, Hyunsung Go, Won-Seok Choi, Kibeom Kim, and Byoung-Tak Zhang.
\newblock Learning geometry-aware representations by sketching.
\newblock \emph{arXiv preprint arXiv:2304.08204}, 2023.

\bibitem[Li et~al.(2018)Li, Fu, and Tai]{li2018fast}
Lei Li, Hongbo Fu, and Chiew-Lan Tai.
\newblock Fast sketch segmentation and labeling with deep learning.
\newblock \emph{IEEE computer graphics and applications}, 39\penalty0 (2):\penalty0 38--51, 2018.

\bibitem[Li et~al.(2023)Li, Wang, Duan, and Li]{li2023clip}
Yi Li, Hualiang Wang, Yiqun Duan, and Xiaomeng Li.
\newblock Clip surgery for better explainability with enhancement in open-vocabulary tasks, 2023.

\bibitem[Lin et~al.(2014)Lin, Maire, Belongie, Hays, Perona, Ramanan, Doll{\'a}r, and Zitnick]{lin2014microsoft}
Tsung-Yi Lin, Michael Maire, Serge Belongie, James Hays, Pietro Perona, Deva Ramanan, Piotr Doll{\'a}r, and C~Lawrence Zitnick.
\newblock Microsoft coco: Common objects in context.
\newblock In \emph{Computer Vision--ECCV 2014: 13th European Conference, Zurich, Switzerland, September 6-12, 2014, Proceedings, Part V 13}, pages 740--755. Springer, 2014.

\bibitem[Liu et~al.(2023)Liu, Zeng, Ren, Li, Zhang, Yang, Li, Yang, Su, Zhu, et~al.]{liu2023grounding}
Shilong Liu, Zhaoyang Zeng, Tianhe Ren, Feng Li, Hao Zhang, Jie Yang, Chunyuan Li, Jianwei Yang, Hang Su, Jun Zhu, et~al.
\newblock Grounding dino: Marrying dino with grounded pre-training for open-set object detection.
\newblock \emph{arXiv preprint arXiv:2303.05499}, 2023.

\bibitem[Long et~al.(2015)Long, Shelhamer, and Darrell]{long2015fully}
Jonathan Long, Evan Shelhamer, and Trevor Darrell.
\newblock Fully convolutional networks for semantic segmentation.
\newblock In \emph{Proceedings of the IEEE conference on computer vision and pattern recognition}, pages 3431--3440, 2015.

\bibitem[L{\"u}ddecke and Ecker(2022)]{luddecke2022image}
Timo L{\"u}ddecke and Alexander Ecker.
\newblock Image segmentation using text and image prompts.
\newblock In \emph{Proceedings of the IEEE/CVF Conference on Computer Vision and Pattern Recognition}, pages 7086--7096, 2022.

\bibitem[Luo et~al.(2022)Luo, Bao, Wu, He, and Li]{luo2022segclip}
Huaishao Luo, Junwei Bao, Youzheng Wu, Xiaodong He, and Tianrui Li.
\newblock Segclip: Patch aggregation with learnable centers for open-vocabulary semantic segmentation.
\newblock \emph{arXiv e-prints}, pages arXiv--2211, 2022.

\bibitem[Melas-Kyriazi et~al.(2022)Melas-Kyriazi, Rupprecht, Laina, and Vedaldi]{melas2022deep}
Luke Melas-Kyriazi, Christian Rupprecht, Iro Laina, and Andrea Vedaldi.
\newblock Deep spectral methods: A surprisingly strong baseline for unsupervised semantic segmentation and localization.
\newblock In \emph{Proceedings of the IEEE/CVF Conference on Computer Vision and Pattern Recognition}, pages 8364--8375, 2022.

\bibitem[Mittal et~al.(2019)Mittal, Tatarchenko, and Brox]{mittal2019semi}
Sudhanshu Mittal, Maxim Tatarchenko, and Thomas Brox.
\newblock Semi-supervised semantic segmentation with high-and low-level consistency.
\newblock \emph{IEEE transactions on pattern analysis and machine intelligence}, 43\penalty0 (4):\penalty0 1369--1379, 2019.

\bibitem[Pathak et~al.(2015)Pathak, Shelhamer, Long, and Darrell]{pathak2014fully}
Deepak Pathak, Evan Shelhamer, Jonathan Long, and Trevor Darrell.
\newblock Fully convolutional multi-class multiple instance learning.
\newblock In \emph{ICLR Workshop}, 2015.

\bibitem[Qi et~al.(2022)Qi, Gryaditskaya, Xiang, and Song]{qi2022one}
Anran Qi, Yulia Gryaditskaya, Tao Xiang, and Yi-Zhe Song.
\newblock One sketch for all: One-shot personalized sketch segmentation.
\newblock \emph{IEEE transactions on image processing}, 31:\penalty0 2673--2682, 2022.

\bibitem[Qi and Tan(2019)]{qi2019sketchsegnet+}
Yonggang Qi and Zheng-Hua Tan.
\newblock Sketchsegnet+: An end-to-end learning of rnn for multi-class sketch semantic segmentation.
\newblock \emph{Ieee Access}, 7:\penalty0 102717--102726, 2019.

\bibitem[Radford et~al.(2021)Radford, Kim, Hallacy, Ramesh, Goh, Agarwal, Sastry, Askell, Mishkin, Clark, et~al.]{radford2021learning}
Alec Radford, Jong~Wook Kim, Chris Hallacy, Aditya Ramesh, Gabriel Goh, Sandhini Agarwal, Girish Sastry, Amanda Askell, Pamela Mishkin, Jack Clark, et~al.
\newblock Learning transferable visual models from natural language supervision.
\newblock In \emph{International conference on machine learning}, pages 8748--8763. PMLR, 2021.

\bibitem[Rao et~al.(2022)Rao, Zhao, Chen, Tang, Zhu, Huang, Zhou, and Lu]{rao2022denseclip}
Yongming Rao, Wenliang Zhao, Guangyi Chen, Yansong Tang, Zheng Zhu, Guan Huang, Jie Zhou, and Jiwen Lu.
\newblock Denseclip: Language-guided dense prediction with context-aware prompting.
\newblock In \emph{Proceedings of the IEEE/CVF Conference on Computer Vision and Pattern Recognition}, pages 18082--18091, 2022.

\bibitem[Sain et~al.(2023)Sain, Bhunia, Chowdhury, Koley, Xiang, and Song]{sain2023clip}
Aneeshan Sain, Ayan~Kumar Bhunia, Pinaki~Nath Chowdhury, Subhadeep Koley, Tao Xiang, and Yi-Zhe Song.
\newblock Clip for all things zero-shot sketch-based image retrieval, fine-grained or not.
\newblock \emph{arXiv preprint arXiv:2303.13440}, 2023.

\bibitem[Sangkloy et~al.(2016)Sangkloy, Burnell, Ham, and Hays]{sangkloy2016sketchy}
Patsorn Sangkloy, Nathan Burnell, Cusuh Ham, and James Hays.
\newblock The sketchy database: learning to retrieve badly drawn bunnies.
\newblock \emph{ACM Transactions on Graphics (TOG)}, 35\penalty0 (4), 2016.

\bibitem[Sangkloy et~al.(2022)Sangkloy, Jitkrittum, Yang, and Hays]{sangkloy2022sketch}
Patsorn Sangkloy, Wittawat Jitkrittum, Diyi Yang, and James Hays.
\newblock A sketch is worth a thousand words: Image retrieval with text and sketch.
\newblock In \emph{Computer Vision--ECCV 2022: 17th European Conference, Tel Aviv, Israel, October 23--27, 2022, Proceedings, Part XXXVIII}, pages 251--267. Springer, 2022.

\bibitem[Scarselli et~al.(2008)Scarselli, Gori, Tsoi, Hagenbuchner, and Monfardini]{scarselli2008graph}
Franco Scarselli, Marco Gori, Ah~Chung Tsoi, Markus Hagenbuchner, and Gabriele Monfardini.
\newblock The graph neural network model.
\newblock \emph{IEEE transactions on neural networks}, 20\penalty0 (1):\penalty0 61--80, 2008.

\bibitem[Schaldenbrand et~al.(2021)Schaldenbrand, Liu, and Oh]{schaldenbrand2021styleclipdraw}
Peter Schaldenbrand, Zhixuan Liu, and Jean Oh.
\newblock Styleclipdraw: Coupling content and style in text-to-drawing synthesis.
\newblock \emph{arXiv preprint arXiv:2111.03133}, 2021.

\bibitem[Schlachter et~al.(2022)Schlachter, Ahlbrand, Wang, Perlin, and Ortenzi]{schlachter2022zero}
Kristofer Schlachter, Benjamin Ahlbrand, Zhu Wang, Ken Perlin, and Valerio Ortenzi.
\newblock Zero-shot multi-modal artist-controlled retrieval and exploration of 3d object sets.
\newblock In \emph{SIGGRAPH Asia 2022 Technical Communications}, pages 1--4. 2022.

\bibitem[Sun et~al.(2012)Sun, Wang, Zhang, and Zhang]{sun2012free}
Zhenbang Sun, Changhu Wang, Liqing Zhang, and Lei Zhang.
\newblock Free hand-drawn sketch segmentation.
\newblock In \emph{Computer Vision--ECCV 2012: 12th European Conference on Computer Vision, Florence, Italy, October 7-13, 2012, Proceedings, Part I 12}, pages 626--639. Springer, 2012.

\bibitem[Touvron et~al.(2021)Touvron, Cord, Douze, Massa, Sablayrolles, and J{\'e}gou]{touvron2021training}
Hugo Touvron, Matthieu Cord, Matthijs Douze, Francisco Massa, Alexandre Sablayrolles, and Herv{\'e} J{\'e}gou.
\newblock Training data-efficient image transformers \& distillation through attention.
\newblock In \emph{International conference on machine learning}, pages 10347--10357. PMLR, 2021.

\bibitem[Vinker et~al.(2022{\natexlab{a}})Vinker, Alaluf, Cohen-Or, and Shamir]{vinker2022clipascene}
Yael Vinker, Yuval Alaluf, Daniel Cohen-Or, and Ariel Shamir.
\newblock Clipascene: Scene sketching with different types and levels of abstraction.
\newblock \emph{arXiv preprint arXiv:2211.17256}, 2022{\natexlab{a}}.

\bibitem[Vinker et~al.(2022{\natexlab{b}})Vinker, Pajouheshgar, Bo, Bachmann, Bermano, Cohen-Or, Zamir, and Shamir]{vinker2022clipasso}
Yael Vinker, Ehsan Pajouheshgar, Jessica~Y Bo, Roman~Christian Bachmann, Amit~Haim Bermano, Daniel Cohen-Or, Amir Zamir, and Ariel Shamir.
\newblock Clipasso: Semantically-aware object sketching.
\newblock \emph{ACM Transactions on Graphics (TOG)}, 41\penalty0 (4):\penalty0 1--11, 2022{\natexlab{b}}.

\bibitem[Wang et~al.(2020)Wang, Lin, Li, Wu, Cai, Luo, and Wang]{wang2020multi}
Fei Wang, Shujin Lin, Hanhui Li, Hefeng Wu, Tie Cai, Xiaonan Luo, and Ruomei Wang.
\newblock Multi-column point-cnn for sketch segmentation.
\newblock \emph{Neurocomputing}, 392:\penalty0 50--59, 2020.

\bibitem[Wei et~al.(2018)Wei, Xiao, Shi, Jie, Feng, and Huang]{wei2018revisiting}
Yunchao Wei, Huaxin Xiao, Honghui Shi, Zequn Jie, Jiashi Feng, and Thomas~S Huang.
\newblock Revisiting dilated convolution: A simple approach for weakly-and semi-supervised semantic segmentation.
\newblock In \emph{Proceedings of the IEEE conference on computer vision and pattern recognition}, 2018.

\bibitem[Wu et~al.(2018)Wu, Qi, Liu, and Yang]{wu2018sketchsegnet}
Xingyuan Wu, Yonggang Qi, Jun Liu, and Jie Yang.
\newblock Sketchsegnet: A rnn model for labeling sketch strokes.
\newblock In \emph{2018 IEEE 28th International Workshop on Machine Learning for Signal Processing (MLSP)}, pages 1--6. IEEE, 2018.

\bibitem[Xu et~al.(2022)Xu, De~Mello, Liu, Byeon, Breuel, Kautz, and Wang]{xu2022groupvit}
Jiarui Xu, Shalini De~Mello, Sifei Liu, Wonmin Byeon, Thomas Breuel, Jan Kautz, and Xiaolong Wang.
\newblock Groupvit: Semantic segmentation emerges from text supervision.
\newblock In \emph{Proceedings of the IEEE/CVF Conference on Computer Vision and Pattern Recognition}, pages 18134--18144, 2022.

\bibitem[Yang et~al.(2023)Yang, Ke, Yu, and Cai]{yang2023scene}
Jie Yang, Aihua Ke, Yaoxiang Yu, and Bo Cai.
\newblock Scene sketch semantic segmentation with hierarchical transformer.
\newblock \emph{Knowledge-Based Systems}, page 110962, 2023.

\bibitem[Yang et~al.(2021)Yang, Zhuang, Fu, Wei, Zhou, and Zheng]{yang2020sketchgcn}
Lumin Yang, Jiajie Zhuang, Hongbo Fu, Xiangzhi Wei, Kun Zhou, and Youyi Zheng.
\newblock Sketchgnn: Semantic sketch segmentation with graph neural networks.
\newblock \emph{ACM Trans. Graph.}, 40\penalty0 (3):\penalty0 1--13, 2021.

\bibitem[Yao et~al.(2022)Yao, Cui, Li, and Gu]{yao2022improving}
Ruichen Yao, Ziteng Cui, Xiaoxiao Li, and Lin Gu.
\newblock Improving fairness in image classification via sketching.
\newblock \emph{arXiv preprint arXiv:2211.00168}, 2022.

\bibitem[Zadaianchuk et~al.(2022)Zadaianchuk, Kleindessner, Zhu, Locatello, and Brox]{zadaianchuk2022unsupervised}
Andrii Zadaianchuk, Matthaeus Kleindessner, Yi Zhu, Francesco Locatello, and Thomas Brox.
\newblock Unsupervised semantic segmentation with self-supervised object-centric representations.
\newblock \emph{arXiv preprint arXiv:2207.05027}, 2022.

\bibitem[Zhang et~al.(2022)Zhang, Deng, Li, Lai, Ma, Liu, and Wang]{zhang2022stroke}
Zhengming Zhang, Xiaoming Deng, Jinyao Li, Yukun Lai, Cuixia Ma, Yongjin Liu, and Hongan Wang.
\newblock Stroke-based semantic segmentation for scene-level free-hand sketches.
\newblock \emph{The Visual Computer}, pages 1--13, 2022.

\bibitem[Zhao et~al.(2017)Zhao, Shi, Qi, Wang, and Jia]{zhao2017pyramid}
Hengshuang Zhao, Jianping Shi, Xiaojuan Qi, Xiaogang Wang, and Jiaya Jia.
\newblock Pyramid scene parsing network.
\newblock In \emph{Proceedings of the IEEE conference on computer vision and pattern recognition}, pages 2881--2890, 2017.

\bibitem[Zheng et~al.(2023{\natexlab{a}})Zheng, Pan, Wang, Tong, Liu, and Shum]{zheng2023locally}
Xin-Yang Zheng, Hao Pan, Peng-Shuai Wang, Xin Tong, Yang Liu, and Heung-Yeung Shum.
\newblock Locally attentional sdf diffusion for controllable 3d shape generation.
\newblock \emph{ACM TOG, Proc. SIGGRAPH}, 2023{\natexlab{a}}.

\bibitem[Zheng et~al.(2023{\natexlab{b}})Zheng, Xie, Sain, Song, and Ma]{zheng2023sketch}
Yixiao Zheng, Jiyang Xie, Aneeshan Sain, Yi-Zhe Song, and Zhanyu Ma.
\newblock Sketch-segformer: Transformer-based segmentation for figurative and creative sketches.
\newblock \emph{IEEE Transactions on Image Processing}, 2023{\natexlab{b}}.

\bibitem[Zhou et~al.(2021)Zhou, Loy, and Dai]{zhou2021denseclip}
Chong Zhou, Chen~Change Loy, and Bo Dai.
\newblock Denseclip: Extract free dense labels from clip.
\newblock \emph{arXiv preprint arXiv:2112.01071}, 2021.

\bibitem[Zhou et~al.(2022)Zhou, Yang, Loy, and Liu]{zhou2022conditional}
Kaiyang Zhou, Jingkang Yang, Chen~Change Loy, and Ziwei Liu.
\newblock Conditional prompt learning for vision-language models.
\newblock In \emph{Proceedings of the IEEE/CVF Conference on Computer Vision and Pattern Recognition}, 2022.

\bibitem[Zhou et~al.(2023)Zhou, Lei, Zhang, Liu, and Liu]{zhou2023zegclip}
Ziqin Zhou, Yinjie Lei, Bowen Zhang, Lingqiao Liu, and Yifan Liu.
\newblock Zegclip: Towards adapting clip for zero-shot semantic segmentation.
\newblock In \emph{Proceedings of the IEEE/CVF Conference on Computer Vision and Pattern Recognition}, pages 11175--11185, 2023.

\bibitem[Zhu et~al.(2018)Zhu, Xiao, and Zheng]{zhu2018part}
Xianyi Zhu, Yi Xiao, and Yan Zheng.
\newblock Part-level sketch segmentation and labeling using dual-cnn.
\newblock In \emph{Neural Information Processing: 25th International Conference, ICONIP 2018, Siem Reap, Cambodia, December 13-16, 2018, Proceedings, Part I 25}, pages 374--384. Springer, 2018.

\bibitem[Zhu et~al.(2020)Zhu, Xiao, and Zheng]{zhu20202d}
Xianyi Zhu, Yi Xiao, and Yan Zheng.
\newblock 2d freehand sketch labeling using cnn and crf.
\newblock \emph{Multimed. Tools. Appl.}, 79\penalty0 (1), 2020.

\bibitem[Zhu et~al.(2021)Zhu, Zhang, Wu, Zhang, He, Zhang, Manmatha, Li, and Smola]{zhu2021improving}
Y Zhu, Z Zhang, C Wu, Z Zhang, T He, H Zhang, R Manmatha, M Li, and A Smola.
\newblock Improving semantic segmentation via self-training. arxiv 2020.
\newblock \emph{arXiv preprint arXiv:2004.14960}, 2021.

\bibitem[Zou et~al.(2018)Zou, Yu, Du, Mo, Song, Xiang, Gao, Chen, and Zhang]{zou2018sketchyscene}
Changqing Zou, Qian Yu, Ruofei Du, Haoran Mo, Yi-Zhe Song, Tao Xiang, Chengying Gao, Baoquan Chen, and Hao Zhang.
\newblock Sketchyscene: Richly-annotated scene sketches.
\newblock In \emph{Proceedings of the european conference on computer vision (ECCV)}, pages 421--436, 2018.

\end{thebibliography}
}

\clearpage
\renewcommand\thefigure{S\arabic{figure}}
\renewcommand\thetable{S\arabic{table}}
\renewcommand\thesection{S\arabic{section}}
\renewcommand\thesubsection{S\arabic{section}.\arabic{subsection}}

\setcounter{section}{0}

\section{Overview of the Supplementary Material}
\begin{itemize}
    \item{}
    In \cref{sec:sup_visual_comparisons}, we provide \textbf{additional visual comparisons} of the results obtained with our method versus results obtained with the state-of-the-art language-supervised image segmentation methods.
    \item{} 
    In \cref{sec_sup:segmentation_accuracy_cat}, we analyze \textbf{segmentation accuracy per category}.
    \item{} 
    In \cref{sec:sup_generalization}, we further investigate \textbf{the generalization properties} of our method and how it compares with fully-supervised methods.
    \item{}
    In \cref{sec:sup_user_study}, we provide a more in-depth discussion of \cref{sec:human_model_align}: \emph{Human-model alignment} of the main paper.
    \item{}
    In \cref{sec:sup_cross_attentio}, we provide a detailed analysis of the \textbf{benefit of using cross-attention}.
    \item{}
    In \cref{sec_sup:checkpoint_choice}, we analyze different models' performance depending on \textbf{the choice of a checkpoint}: the last checkpoint versus the checkpoint optimal on the validation set.  
    \item{}
    In \cref{sec_sup:thresholding}, we discuss in detail \textbf{the choice of a threshold value} for segmenting out pixels corresponding to \textbf{individual categories}. 
    \item{}
    In \cref{sec_sup:Computational Cost}, we provide the \textbf{computational cost} of our method.
\end{itemize}

% \section{Additional Analysis of our method performance}
\section{Additional Performance Analysis}

\subsection{Additional Qualitative Comparisons}
\label{sec:sup_visual_comparisons}
\begin{figure*}[ht]
  \centering
   \includegraphics[width=1.0\linewidth]{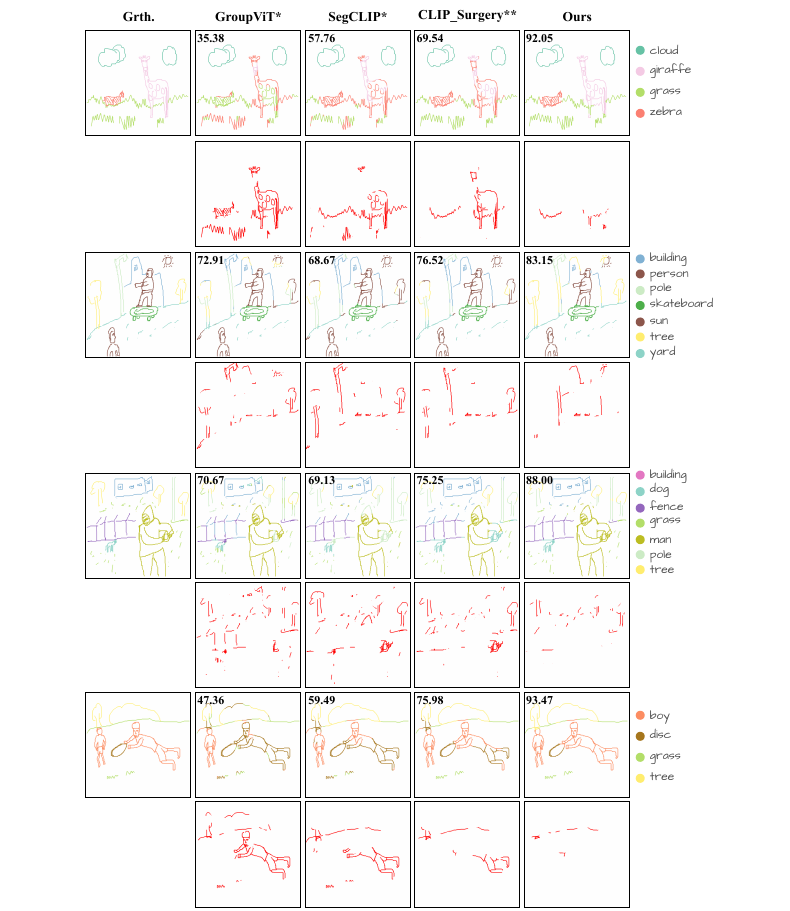}
   \caption{\textbf{Part-1}: Visual comparison of our method against state-of-the-art language supervised image segmentation methods, trained on the FS-COCO dataset \cite{chowdhury2022fs}. 
   The numbers show Acc@P values. 
   The error maps in red represent the misclassified pixels. }
   \label{fig:seg_results}
\end{figure*}

\begin{figure*}[ht]
  \centering
   \includegraphics[width=1.0\linewidth]{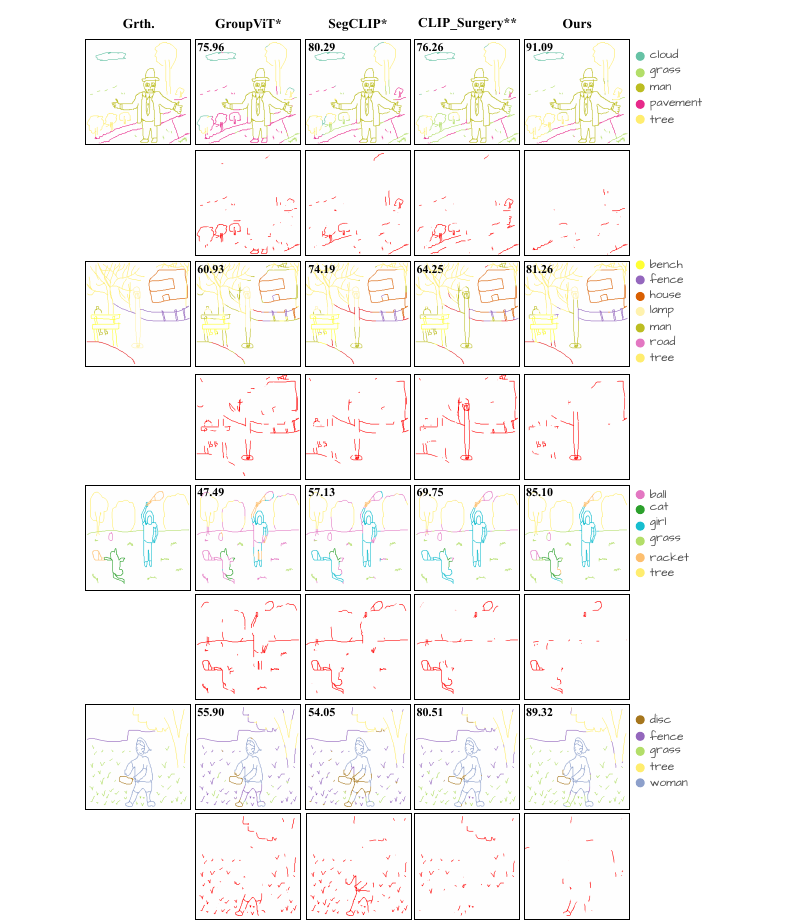}
   \caption{\textbf{Part-2}: Visual comparison of our method against state-of-the-art language supervised image segmentation methods, trained on the FS-COCO dataset \cite{chowdhury2022fs}. 
   The numbers show Acc@P values. 
   The error maps in red represent the misclassified pixels. }   
   \label{fig:seg_results_2}
\end{figure*}

In the main paper, we show in \cref{tab:results_main} a numerical comparison of the segmentation results obtained with our method and the segmentation results obtained with the state-of-the-art language-supervised image segmentation methods. 
Also, in the main paper, in \cref{fig:r1}, we show a comparison of our model with CLIP\_Surgery\textsuperscript{$\star$$\star$}, where \emph{CLIP Surgery\textsuperscript{$\star$$\star$}} represents the fine-tuned  CLIP\_Surgery \cite{li2023clip} model with v-v self-attention introduced at both training and inference stages.
Here, in \cref{fig:seg_results,fig:seg_results_2}, we provide an additional visual comparison between our method and state-of-the-art language-supervised image segmentation methods: GroupViT \cite{xu2022groupvit}, SegCLIP \cite{luo2022segclip}, CLIP\_Surgery \cite{li2023clip}, fine-tuned on the FS-COCO dataset. 
The fine-tuned versions of these models are denoted as GroupViT\textsuperscript{$\star$$\star$}, SegCLIP\textsuperscript{$\star$}, CLIP\_Surgery\textsuperscript{$\star$$\star$}, respectively.
In \cref{fig:seg_results,fig:seg_results_2}, we show segmentation results and the error maps (in red), which visualize incorrectly labeled pixels for each method.

%%%%%%%%%%%%%%%%%%%%%%%%%%%%%%%%%%%%%%%%%%%%%%
% Segmentation accuracy analysis by category
%%%%%%%%%%%%%%%%%%%%%%%%%%%%%%%%%%%%%%%%%%%%%%

\begin{figure*}[ht]
  \centering
   \includegraphics[width=\linewidth]{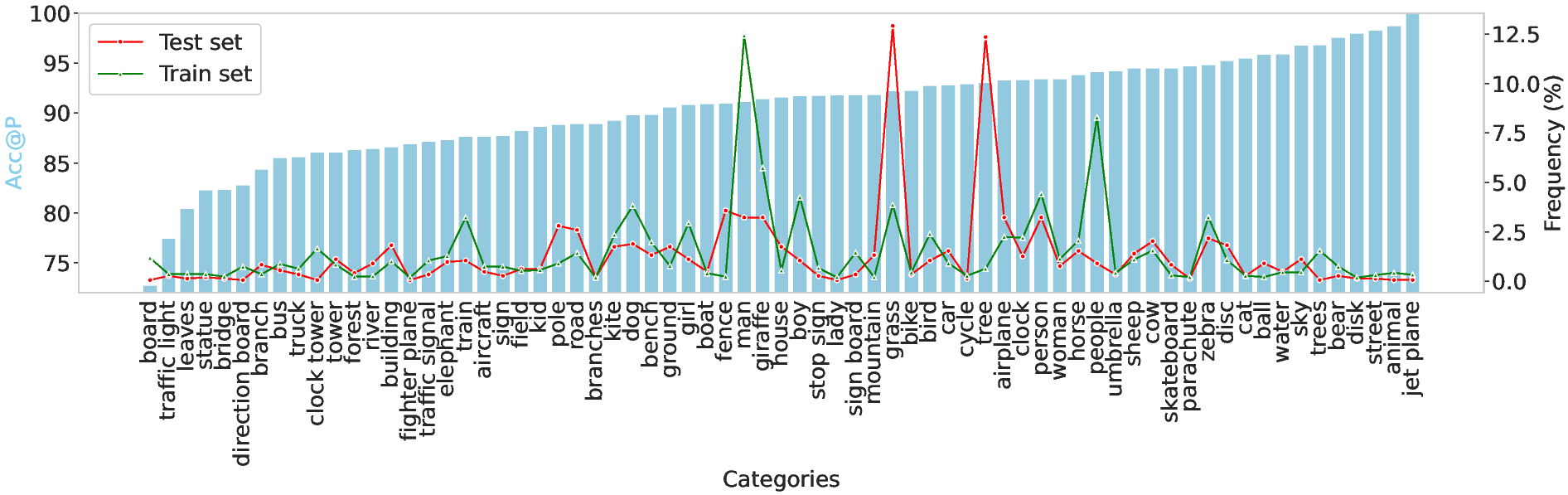}
   \caption{Blue bars show pixel accuracy (Acc@P) for each object category with more than 10 appearances in FS-COCO dataset \cite{chowdhury2022fs} captions.
    The green line shows the frequency of occurrence of each category in the train set.
   The red line shows the frequency of occurrence of each category in the test set.
   Please see \cref{sec_sup:segmentation_accuracy_cat} for an additional discussion.}
   \label{fig:r4}
\end{figure*}

\subsection{Segmentation Accuracy Analysis by Category}
\label{sec_sup:segmentation_accuracy_cat}
In this section, we analyze segmentation accuracy \emph{per category in both the train and test sets}.
We show in \cref{fig:r4} the pixel accuracy (\emph{Acc@P}) for each selected object category.
For the figure, we selected categories that appear more than ten times in the FS-COCO dataset \cite{chowdhury2022fs} captions.
First, we can see that the segmentation accuracy is smoothly distributed across different categories.

Next, we investigate whether more frequent categories are more likely to be labeled accurately. 
To evaluate this, we approximate the frequency of a category by counting its occurrence in both the train and test sets, then consider only categories that appear in the test set. 
We plot with green and red lines in \cref{fig:r4} the train and test sets category frequency, respectively.

The figure clearly shows a lack of correlation between the frequency of category occurrence and its segmentation accuracy. 

We further evaluate it numerically by computing the correlation between $x$, the pixel accuracy (Acc@P) of each category, and $y$ the occurrence frequency of this category:
\begin{equation}
\label{eq:correlation}
    Corr = \frac{N(\sum xy) - (\sum x)(\sum y)}{\sqrt{[N\sum x^2 - (\sum x)^2][N\sum y^2 - (\sum y)^2]}}
\end{equation}
where $N$ is the number of categories in the test set.

The resulting correlation coefficients for both train and test sets are 0.16 and 0.14, respectively. This suggests a very weak accuracy-frequency correspondence, indicating that our model is not biased toward more frequently occurring categories. 
We hypothesize that this is in part due to our careful fine-tuning strategy, which prevents over-fitting. 
Therefore, the model efficiently leverages pre-training on a large image dataset.

\paragraph{Model generalization to new object categories}  
Our test set includes 185 object classes, with 125 seen and 60 unseen during training.
The accuracy on seen categories is $86.35\%$ and $84.68\%$ on unseen.
These results demonstrate \emph{good generalization} of our model to unseen categories. 

% \subsection{Additional Discussion on synthetic vs.~freehand sketches}
\subsection{Synthetic vs.~Freehand sketches}

\label{sec:sup_generalization}
In the \cref{sec:generalization} in the main paper, to better understand the generalization properties of our model, we evaluated our method trained on the sketches from the FS-COCO dataset \cite{chowdhury2022fs} on the freehand sketches from \cite{ge2022exploring}.
Here, we provide additional analysis of generalization properties.

\subsubsection{Generalization to sketches consisting of clip-art-like object sketches}
Here, we additionally evaluate our method on the SketchyScene \cite{zou2018sketchyscene} dataset. 
The SketchyScene \cite{zou2018sketchyscene}  dataset contains 7,264 sketch-image pairs. 
It is obtained by providing participants with a reference image and
clip-art-like object sketches to drag and drop for scene composition. 
The augmentation is performed by replacing object sketches with other sketch instances belonging to the same object category.
This is a dataset with sketches with a large domain gap from the freehand scene sketches we target. 
Yet, it is interesting to evaluate the generalization properties of our method.
\cref{tab:sup_zs_synthetic} shows a comparison of the zero-shot performance of our method (\emph{third line: Ours}) with the two fully-supervised methods trained on semi-synthetic sketches. 
The \emph{Acc@P} and \emph{mIoU} are the metrics we use in the main paper.
We additionally report results for two additional measures:
\begin{itemize}
    \item{\textbf{Mean Pixel Accuracy (MeanAcc)}:} It measures the average pixel accuracy Acc@P of each category.
    \item{\textbf{Frequency Weighted Intersection over Union (FWIoU):}} It introduces category occurrence frequency to the mIoU, by weighting per-category pixel IoU (intersection over union) by the frequency of occurrence. 
\end{itemize}

\emph{Our model reaches high accuracy on these sketches, even in the presence of a large domain gap.} 
In particular, the performance of our model on these sketches is higher than on the freehand and more challenging sketches from the FS-COCO dataset \cite{chowdhury2022fs}.
This, combined with the results in \cref{tab:zs_freehand} in the main paper, is a strong argument towards usage of \emph{true} freehand sketches with weak annotation in the form of captions over the semi-synthetic dataset of scene sketches.

\begin{table}[t]
  \centering
  \begin{tabular}{@{}lcccc@{}}
    \toprule
    Method & $Acc@P$ & $MAcc$ & $mIoU$&$FWIoU$    \\
    \midrule
    LDP \cite{ge2022exploring} & $93.46$ & $85.84$ & $74.93 $ & $88.13$\\
    SketchSeger \cite{yang2023scene} & $95.44$ & $88.18$ & $81.17$ & $91.52$\\
    Ours & $87.99$ & $66.59$ & $60.91$ & $76.33$\\
    \midrule
    Ours\textsuperscript{$\star$} & $92.87$ & $79.54$ & $71.73$ & $85.19$\\
    Ours\textsuperscript{$\star\star$} & $91.23$ & $77.87$ & $70.51$ & $84.72$\\
    \bottomrule
  \end{tabular}
  \caption{Comparison of our method with state-of-the-art fully supervised scene sketch segmentation methods on the sketches from the SketchyScene \cite{zou2018sketchyscene} dataset.
     \emph{Ours}: trained on freehand sketches from the FS-COCO dataset \cite{chowdhury2022fs} (zero-shot performance), \emph{Ours\textsuperscript{$\star$}} is trained on synthetic sketches \cite{zou2018sketchyscene}, \emph{Ours\textsuperscript{$\star\star$}} is trained on both freehand \cite{chowdhury2022fs}  and synthetic sketches \cite{zou2018sketchyscene}.}
  \label{tab:sup_zs_synthetic}
\end{table}

\begin{table*}[t]
\vspace{1.2cm}
\resizebox{\textwidth}{!}{%
\begin{tabular}{|c|c|ll|c|ccccc|}
\multirow{2}{*}{Method}     & \multirow{2}{*}{Trained on}        & \multicolumn{2}{c|}{Supervision}                                             & \multirow{2}{*}{Tested on} & \multicolumn{5}{c|}{Segmentation accuracy}                                                                                                 \\
                             &                               & \multicolumn{1}{c}{Pixel labels}      & \multicolumn{1}{c|}{Captions} &                           & \multicolumn{1}{c|}{mIoU}  & \multicolumn{1}{c|}{Acc@P} & \multicolumn{1}{c|}{Acc@C} & \multicolumn{1}{c|}{MAcc}  & FWIoU \\
\cmidrule(lr){1-5} \cmidrule(lr){6-10} 
\multirow{3}{*}{LDP \cite{ge2022exploring}}         & SketchyScene $\cup$                  & \multicolumn{1}{l}{}                  & \multirow{3}{*}{}                   & \cellcolor[HTML]{FFF2CC}FS-COCO                   & \multicolumn{1}{c|}{\cellcolor[HTML]{FFF2CC}33.04} & \multicolumn{1}{c|}{\cellcolor[HTML]{FFF2CC}56.23} & \multicolumn{1}{c|}{\cellcolor[HTML]{FFF2CC}56.71} & \multicolumn{1}{c|}{\cellcolor[HTML]{FFF2CC}51.16} & \cellcolor[HTML]{FFF2CC}52.63 \\
                             & $\cup$  SKY-Scene   $\cup$                   & \multicolumn{1}{c}{\checkmark}               &                                     & \cellcolor[HTML]{DAE8FC}LDP freehand              & \multicolumn{1}{c|}{\cellcolor[HTML]{DAE8FC}37.16} & \multicolumn{1}{c|}{\cellcolor[HTML]{DAE8FC}78.84} & \multicolumn{1}{c|}{\cellcolor[HTML]{DAE8FC}-}     & \multicolumn{1}{c|}{\cellcolor[HTML]{DAE8FC}47.25} & \cellcolor[HTML]{DAE8FC}66.98 \\
                             & $\cup$  TUB-Scene                     & \multicolumn{1}{l}{}                  &                                     & SketchyScene            & \multicolumn{1}{c|}{74.93} & \multicolumn{1}{c|}{93.46} & \multicolumn{1}{c|}{-}     & \multicolumn{1}{c|}{85.84} & 88.13 \\
\cmidrule(lr){1-5} \cmidrule(lr){6-10} 
\multirow{3}{*}{SketchSeger \cite{yang2023scene}} & SketchyScene $\cup$                  & \multicolumn{1}{l}{}                  & \multirow{3}{*}{}                   & \cellcolor[HTML]{FFF2CC}FS-COCO                   & \multicolumn{1}{c|}{\cellcolor[HTML]{FFF2CC}-}     & \multicolumn{1}{c|}{\cellcolor[HTML]{FFF2CC}-}     & \multicolumn{1}{c|}{\cellcolor[HTML]{FFF2CC}-}     & \multicolumn{1}{c|}{\cellcolor[HTML]{FFF2CC}-}     & \cellcolor[HTML]{FFF2CC}-     \\
                             & $\cup$  SKY-Scene $\cup$                      & \multicolumn{1}{c}{\checkmark}               &                                     & \cellcolor[HTML]{DAE8FC}LDP freehand              & \multicolumn{1}{c|}{\cellcolor[HTML]{DAE8FC}-}     & \multicolumn{1}{c|}{\cellcolor[HTML]{DAE8FC}-}     & \multicolumn{1}{c|}{\cellcolor[HTML]{DAE8FC}-}     & \multicolumn{1}{c|}{\cellcolor[HTML]{DAE8FC}-}     & \cellcolor[HTML]{DAE8FC}-     \\
                             & $\cup$  TUB-Scene                     & \multicolumn{1}{l}{}                  &                                     & SketchyScene               & \multicolumn{1}{c|}{81.17} & \multicolumn{1}{c|}{95.44} & \multicolumn{1}{c|}{-}     & \multicolumn{1}{c|}{88.18} & 91.52 \\
\cmidrule(lr){1-5} \cmidrule(lr){6-10} 
\multirow{3}{*}{Ours}        & \multicolumn{1}{l|}{}         & \multicolumn{1}{l}{\multirow{3}{*}{}} &                                     & \cellcolor[HTML]{FFF2CC}FS-COCO                   & \multicolumn{1}{c|}{\cellcolor[HTML]{FFF2CC}73.48} & \multicolumn{1}{c|}{\cellcolor[HTML]{FFF2CC}85.54} & \multicolumn{1}{c|}{\cellcolor[HTML]{FFF2CC}87.02} & \multicolumn{1}{c|}{\cellcolor[HTML]{FFF2CC}82.27} & \cellcolor[HTML]{FFF2CC}84.09 \\
                             & FS-COCO                       & \multicolumn{1}{l}{}                  & \multicolumn{1}{c|}{\checkmark}            & \cellcolor[HTML]{DAE8FC}LDP freehand              & \multicolumn{1}{c|}{\cellcolor[HTML]{DAE8FC}53.94} & \multicolumn{1}{c|}{\cellcolor[HTML]{DAE8FC}81.63} & \multicolumn{1}{c|}{\cellcolor[HTML]{DAE8FC}-}     & \multicolumn{1}{c|}{\cellcolor[HTML]{DAE8FC}59.36} & \cellcolor[HTML]{DAE8FC}69.37 \\
                             & \multicolumn{1}{l|}{}         & \multicolumn{1}{l}{}                  &                                     & SketchyScene               & \multicolumn{1}{c|}{60.91} & \multicolumn{1}{c|}{87.99} & \multicolumn{1}{c|}{-}     & \multicolumn{1}{c|}{66.59} & 76.33 \\
\cmidrule(lr){1-5} \cmidrule(lr){6-10} 
\multirow{3}{*}{Ours\textsuperscript{$\star$}}       & \multirow{3}{*}{SketchyScene} & \multicolumn{1}{l}{\multirow{3}{*}{}} &                                     & \cellcolor[HTML]{FFF2CC}FS-COCO                   & \multicolumn{1}{c|}{\cellcolor[HTML]{FFF2CC}61.79} & \multicolumn{1}{c|}{\cellcolor[HTML]{FFF2CC}74.43} & \multicolumn{1}{c|}{\cellcolor[HTML]{FFF2CC}75.62} & \multicolumn{1}{c|}{\cellcolor[HTML]{FFF2CC}69.41} & \cellcolor[HTML]{FFF2CC}71.75 \\
                             &                               & \multicolumn{1}{l}{}                  & \multicolumn{1}{c|}{\checkmark}            & \cellcolor[HTML]{DAE8FC}LDP freehand              & \multicolumn{1}{c|}{\cellcolor[HTML]{DAE8FC}49.72} & \multicolumn{1}{c|}{\cellcolor[HTML]{DAE8FC}71.96} & \multicolumn{1}{c|}{\cellcolor[HTML]{DAE8FC}-}     & \multicolumn{1}{c|}{\cellcolor[HTML]{DAE8FC}48.71} & \cellcolor[HTML]{DAE8FC}59.15 \\
                             &                               & \multicolumn{1}{l}{}                  &                                     & SketchyScene               & \multicolumn{1}{c|}{71.73} & \multicolumn{1}{c|}{92.87} & \multicolumn{1}{c|}{-}     & \multicolumn{1}{c|}{79.54} & 85.19 \\
\cmidrule(lr){1-5} \cmidrule(lr){6-10} 
\multirow{3}{*}{Ours\textsuperscript{$\star\star$}}      & FS-COCO $\cup$                       & \multicolumn{1}{l}{\multirow{3}{*}{}} &                                     & \cellcolor[HTML]{FFF2CC}FS-COCO                   & \multicolumn{1}{c|}{\cellcolor[HTML]{FFF2CC}68.84} & \multicolumn{1}{c|}{\cellcolor[HTML]{FFF2CC}79.21} & \multicolumn{1}{c|}{\cellcolor[HTML]{FFF2CC}81.29} & \multicolumn{1}{c|}{\cellcolor[HTML]{FFF2CC}74.08} & \cellcolor[HTML]{FFF2CC}77.63 \\
                             & $\cup$ SketchyScene                   & \multicolumn{1}{l}{}                  & \multicolumn{1}{c|}{\checkmark}            & \cellcolor[HTML]{DAE8FC}LDP freehand              & \multicolumn{1}{c|}{\cellcolor[HTML]{DAE8FC}50.13} & \multicolumn{1}{c|}{\cellcolor[HTML]{DAE8FC}76.07} & \multicolumn{1}{c|}{\cellcolor[HTML]{DAE8FC}-}     & \multicolumn{1}{c|}{\cellcolor[HTML]{DAE8FC}55.83} & \cellcolor[HTML]{DAE8FC}62.97 \\
                             & \multicolumn{1}{l|}{}         & \multicolumn{1}{l}{}                  &                                     &  SketchyScene               & \multicolumn{1}{c|}{70.51} & \multicolumn{1}{c|}{91.23} & \multicolumn{1}{c|}{-}     & \multicolumn{1}{c|}{77.87} & 84.72
\end{tabular}
}
\vspace{-0.2cm}
\caption{Comparison of our method with state-of-the-art fully supervised scene sketch segmentation methods in different setups. 
\\
\emph{Ours}: trained on freehand sketches from the FS-COCO dataset \cite{chowdhury2022fs}, \emph{Ours\textsuperscript{$\star$}} is trained on synthetic sketches \cite{zou2018sketchyscene}, \emph{Ours\textsuperscript{$\star\star$}} is trained on both freehand \cite{chowdhury2022fs}  and synthetic sketches \cite{zou2018sketchyscene}.
\\
We test all methods on three datasets: 
our FS-COCO-based test set, 
LDP \cite{ge2022exploring} freehand sketches test set, 
and SketchyScene \cite{zou2018sketchyscene} synthetic sketches test set. 
\\
Training datasets: The SketchyScene \cite{zou2018sketchyscene} dataset contains 7,265 synthetic scene sketches spanning 46 categories with 5,617 images for training, and
1,113 for test. SKY-Scene and TUB-Scene were introduced in \cite{ge2022exploring}, and are composed of object sketches from the Sketchy \cite{sangkloy2016sketchy} and TU-Berlin \cite{eitz2012humans} datasets, respectively. 
They both have 7,265 synthetic scene sketches and follow the same data split.}
\label{tab:sup_results_extensive}
\end{table*}

\begin{table}[h]
  \centering
  \begin{tabular}{@{}lccc@{}}
    \toprule
    Method & $mIoU$ & $Acc@P$ & $Acc@C$   \\
    \midrule
    LDP \cite{ge2022exploring} & $ 33.04$ & $ 56.23$ & $ 56.71$ \\
    Ours & $73.48$ & $85.54$ & $87.02$ \\
    \midrule
    Ours\textsuperscript{$\star$} & $61.79$ & $74.43$ & $ 75.62$ \\
    Ours\textsuperscript{$\star\star$}& $68.84$ & $79.21$ & $81.29$ \\
    \bottomrule
  \end{tabular}
  \caption{Comparison on the freehand sketches from the FS-COCO  dataset \cite{chowdhury2022fs} of our method with state-of-the-art fully supervised scene sketch segmentation method LDP \cite{ge2022exploring}.
    LDP \cite{ge2022exploring} is trained on semi-synthetic sketches.
     \emph{Ours}: trained on freehand sketches from the FS-COCO dataset \cite{chowdhury2022fs}, \emph{Ours\textsuperscript{$\star$}} is trained on synthetic sketches \cite{zou2018sketchyscene}, \emph{Ours\textsuperscript{$\star\star$}} is trained on both freehand \cite{chowdhury2022fs}  and synthetic sketches \cite{zou2018sketchyscene}.
    \emph{We do not compare here with SketchSeger \cite{yang2023scene}, as there is no code available and we can not run it on sketches from the FS-COCO dataset \cite{chowdhury2022fs}.}
  }
  \label{tab:sup_fscoco}
\end{table}

\paragraph{Fine-tuning on semi-synthetic sketches}
While our model does reach high accuracy on these sketches, it does not reach the performance of fully supervised methods trained on semi-synthetic sketches when tested on semi-synthetic sketches.
Therefore, we investigate whether fine-tuning our model on 
semi-synthetic sketches can close the gap -- while relying only on textual labels and not pixel-level annotations. 

We perform two additional experiments: 
\begin{enumerate}
    \item{\textbf{Training exclusively on Synthetic Sketches (Ours\textsuperscript{$\star$}):}}  
    We train our model on the SketchyScene synthetic sketches \cite{zou2018sketchyscene} using language supervision. 
    Captions are constructed by concatenating scene sketch category names into one text token. 
    
    \item{\textbf{Training on Both Synthetic and Freehand Sketches (Ours\textsuperscript{$\star\star$}):}}  We train the model on both SketchyScene synthetic sketches and FS-COCO freehand sketches. 
\end{enumerate}

The results are shown in \cref{tab:sup_zs_synthetic}:  Ours\textsuperscript{$\star$} and Ours\textsuperscript{$\star\star$}.

We observe a performance increase for \emph{Ours\textsuperscript{$\star$}} on the sketches from the SketchyScene \cite{zou2018sketchyscene} dataset, reaching competitive performance with fully supervised methods \cite{ge2022exploring, yang2023scene}. 
\emph{This highlights the generalization properties of our training pipeline for different data distributions and highlights that succinct captions can serve as a robust supervisory signal, lifting the need for extensive annotations.}

However, when freehand sketches are added to the training data (\emph{Ours\textsuperscript{$\star\star$}}), there is a slight decrease in performance across all metrics.
\emph{This further emphasizes the existence of a domain gap between freehand sketches and semi-synthetic sketches, which again motivates the usage of freehand sketches with weak annotations.}

Similar observations are made in \cref{tab:sup_fscoco} when the model is trained on the synthetic sketches (\emph{Ours\textsuperscript{$\star$}}) and tested on the FS-COCO freehand sketches. 
Even when both synthetic and freehand sketches are used for training (\emph{Ours\textsuperscript{$\star\star$}}), the model's performance degrades compared to training solely on freehand sketches. 
This further emphasizes our observations regarding the domain gap between synthetic and freehand sketches.

\cref{tab:sup_results_extensive} shows a full comparison of our method against fully supervised sketch segmentation methods: LDP \cite{ge2022exploring} and SketchSeger \cite{yang2023scene}, across the free datasets: FS-COCO-based test set, 
LDP \cite{ge2022exploring} freehand sketches test set, 
and SketchyScene \cite{zou2018sketchyscene} synthetic sketches test set. 
It shows the superiority of our method on both datasets of freehand scene sketches.

\subsubsection{Pre-training on synthetic sketches}
We also experiment with fine-tuning CLIP and CLIPSurgery on synthetic sketches.
However, training on millions of synthetic sketches is out of the scope of this work due to computational constraints.  
As a feasible experiment, we generated $9025$ synthetic sketches for the reference images in our training set, using \cite{chan2022drawings}, in `contour' style (as the closest to the test set sketches style).
This is the number of sketches identical to the number of sketches we use to train our model.
The accuracy on our test set of fine-tuned this way CLIP and CLIPSurgery increases by negligible $2$ to $3$ points compared to their zero-shot performance.
In comparison, our model outperforms their zero-shot performance by $56.72\%$ and $13.07\%$ points, respectively. 
Training our model from CLIP weights pre-trained on synthetic sketches boosts the performance only by $0.42$ points.

% \clearpage

\section{Detailed human Study Analysis}
\label{sec:sup_user_study}
In this section, we provide a more in-depth discussion of \cref{sec:human_model_align}: \emph{Human-model alignment} of the main paper.

\subsection{Human Study Categories}
\label{sec_sup:study_categories}
\begin{figure}[t]
  \centering
    \hspace{-0.6cm}
   \includegraphics[width=1.05\linewidth]{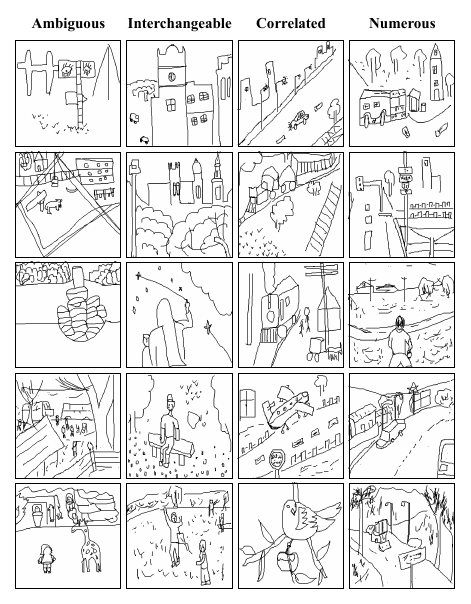}
   \vspace{-0.5cm}
   \caption{Visualization of the selected sketches for the four challenging sketch categories used in the user study. 
Please see \cref{sec_sup:study_categories} for the description of categories.}
   \label{fig:study_categories_2}
\end{figure}

In the main paper, in \cref{sec:challenges}, we introduced four challenging categories of sketches for our method, that we used for the user study.
We show all the sketches used in the user study in \cref{fig:study_categories_2}.
For convenience, below we repeat the definition of each category:
\begin{enumerate}
    \item[(1)]{\textbf{Ambiguous sketches}:} 
    sketches where it might be hard even for a human observer to understand an input sketch. 
    We selected the sketches by visually examining the test set sketches alongside reference images. 
   
   \item[(2)]{\textbf{Interchangeable categories}}: sketches containing multiple objects with labels that can interchange each other, such as \emph{`tower/building'}, \emph{`girl/man'}, and \emph{`ground/grass'}.
   
    \item[(2)]{\textbf{Correlated categories}:} sketches with categories that typically co-occur in scenes. These categories are semantically related. 
    We selected sketches containing the most common pairs with significant co-occurrence. 
    Specifically, \emph{`branch/bird'} (52\%), \emph{`runway/airplane'} (44\%), \emph{`railway/train'} (39\%), and \emph{`road/car'} (29\%), were chosen.

    \item[(4)]{\textbf{Numerous-categories}:} sketches with six or more object categories and a model accuracy (\emph{Acc@P}) below $80\%$. 
    The sampled sketches have an average of $6.4$ categories per sketch $(7,7,6,6,6)$.
\end{enumerate}
Additionally, we included a \textbf{Strong performance} category, comprising ten sketches where the model's accuracy (\emph{Acc@P}) exceeded the average performance (85.54\%), to demonstrate scenarios of effective model segmentation.

\subsection{Annotators}
We recruited 25 participants ($14$ male). 
The annotators are PhD students in diverse disciplines and of diverse nationalities aged from 22 to 42 years (average age 29.32).
% The annotators are 25 (14 male) PhD students in diverse disciplines and of diverse backgrounds basic drawing skills. These annotators' age varies between 22 and 42 years old (average age 29.32). 
We believe this group represents well the general population and each individual performed the task carefully.

% \subsection{Visual Analysis of the segmentation results for the sketches with ``Interchangeable categories" obtained with our model and from the user study}
\subsection{Visual Analysis of Interchangeable Categories Segmentation Results}
\label{sec_sup:human_conf}

\begin{figure}[t]
  \centering
   \includegraphics[width=\linewidth]{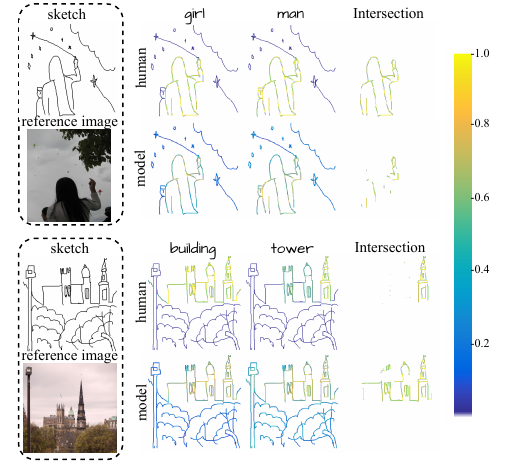}
   \vspace{-0.7cm}
   \caption{Visualizations of the confidence in segmenting semantically similar objects by human annotators and our model. 
   \emph{Intersection} shows the pixels that are confidently assigned to belong to both considered categories (with a confidence threshold higher than $60\%$).
    Please see \cref{sec_sup:human_conf} for the discussion.}
   \label{fig:model_human_confusion}
\end{figure}

We conducted a visual analysis to compare the confidence in segmenting semantically similar objects by human annotators and our model. 
For each object category, we obtain a category confidence map by counting how many participants assigned a given label to a category. 
For our model, we obtain segmentation confidence as a result of a cosine similarity computation between the sketch patch features and the category textual embedding. 
We visualized in \cref{fig:model_human_confusion} the obtained confidence maps for the most frequently confused by our model categories: `girl/man' and `building/tower'.
We also show the pixels that are confidently assigned to belong to both considered categories (with a confidence threshold higher than $60\%$).
We can observe that our model is less confident than humans in assigning labels to these categories.

\subsection{Statistical Significance: Ours vs CLIPSurgery}
On the 20 sketches from the 4 challenging groups, our model outperforms CLIPSurgery with a p-value of $2 \times 10^{-5}$.  
In the `strong' group, we have 10 sketches, in which, while our model performs on par with humans, it outperforms CLIPSurgery with a p-value of $0.005$.

% \clearpage

\section{Additional Ablation Studies}
\subsection{Detailed Ablation on Cross Attention vs.~Self Attention }
\label{sec:sup_cross_attentio}
To validate the effectiveness of our cross-attention module, we added a residual connection to demonstrate that relying solely on self-attention features, without the integration of cross-attention, leads to suboptimal segmentation results.
We run several experiments with varying dropout ratios in the cross-attention block. 
This allows us to assess its impact on model performance. 
The results, presented in \cref{tab:sup_CA_ablation}, show model accuracy across different dropout levels, from 0 (no dropout) to 1 (complete dropout).
This shows the benefit of the design used in the main paper, equivalent to using only cross-attention in the category-level encoder. 
\begin{table}[t]
  \centering
  \resizebox{\columnwidth}{!}{%
  \begin{tabular}{@{}lcccccc@{}}
    \toprule
     Dropout & 0 & 0.2 & 0.4 & 0.6& 0.8 &1    \\
    \midrule
    Acc@P & 85.54 & 84.61 & 84.38& 84.16 & 83.06 & 82.86\\
    \bottomrule
  \end{tabular}
  }
  \caption{Acc@P with different cross attention dropout ratios}
  \label{tab:sup_CA_ablation}
\end{table}

\begin{table*}[!ht]
\centering
\begin{tabular}{l|l|rrr|rrr}
\hline
                                  &                                       & \multicolumn{3}{c|}{\textbf{Test set}}                                                        & \multicolumn{3}{c}{\textbf{Validation set}}                                                   \\
\multirow{-2}{*}{\textbf{Model}}  & \multirow{-2}{*}{\textbf{Checkpoint}} & \multicolumn{1}{c}{mIoU}      & \multicolumn{1}{c}{Acc@P}     & \multicolumn{1}{c|}{Acc@S}    & \multicolumn{1}{c}{mIoU}      & \multicolumn{1}{c}{Acc@P}     & \multicolumn{1}{c}{Acc@S}     \\ \hline
                                  & \cellcolor[HTML]{D8E6D7}Optimal       & \cellcolor[HTML]{D8E6D7}22.86 & \cellcolor[HTML]{D8E6D7}33.41 & \cellcolor[HTML]{D8E6D7}32.64 & \cellcolor[HTML]{D8E6D7}25.76 & \cellcolor[HTML]{D8E6D7}36.34 & \cellcolor[HTML]{D8E6D7}35.17 \\
\multirow{-2}{*}{CLIP*}           & Last                                  & 19.34                         & 28.89                         & 27.64                         & 22.11                         & 31.49                         & 31.07                         \\ \hline
                                  & \cellcolor[HTML]{D8E6D7}Optimal       & \cellcolor[HTML]{D8E6D7}45.71 & \cellcolor[HTML]{D8E6D7}66.21 & \cellcolor[HTML]{D8E6D7}66.89 & \cellcolor[HTML]{D8E6D7}47.26 & \cellcolor[HTML]{D8E6D7}68.28 & \cellcolor[HTML]{D8E6D7}68.76 \\
\multirow{-2}{*}{GroupViT*}       & Last                                  & 43.83                         & 64.03                         & 64.48                         & 46.58                         & 67.70                         & 68.13                         \\ \hline
                                  & \cellcolor[HTML]{D8E6D7}Optimal       & \cellcolor[HTML]{D8E6D7}49.26 & \cellcolor[HTML]{D8E6D7}69.87 & \cellcolor[HTML]{D8E6D7}73.64 & \cellcolor[HTML]{D8E6D7}51.27 & \cellcolor[HTML]{D8E6D7}71.79 & \cellcolor[HTML]{D8E6D7}75.67 \\
\multirow{-2}{*}{SegCLIP*}        & Last                                  & 46.41                         & 66.91                         & 70.31                         & 50.86                         & 70.12                         & 74.41                         \\ \hline
                                  & \cellcolor[HTML]{D8E6D7}Optimal       & \cellcolor[HTML]{D8E6D7}48.74 & \cellcolor[HTML]{D8E6D7}65.38 & \cellcolor[HTML]{D8E6D7}68.78 & \cellcolor[HTML]{D8E6D7}50.84 & \cellcolor[HTML]{D8E6D7}67.32 & \cellcolor[HTML]{D8E6D7}70.88 \\
\multirow{-2}{*}{CLIP\_Surgery*}  & Last                                  & 47.29                         & 63.94                         & 67.13                         & 48.33                         & 66.01                         & 68.82                         \\ \hline
                                  & \cellcolor[HTML]{D8E6D7}Optimal       & \cellcolor[HTML]{D8E6D7}59.98 & \cellcolor[HTML]{D8E6D7}78.68 & \cellcolor[HTML]{D8E6D7}81.11 & \cellcolor[HTML]{D8E6D7}62.41 & \cellcolor[HTML]{D8E6D7}80.69 & \cellcolor[HTML]{D8E6D7}83.23 \\
\multirow{-2}{*}{CLIP\_Surgery**} & Last                                  & 58.64                         & 77.34                         & 79.88                         & 61.53                         & 79.41                         & 82.07                         \\ \hline
                                  & \cellcolor[HTML]{D8E6D7}Optimal       & \cellcolor[HTML]{D8E6D7}73.48 & \cellcolor[HTML]{D8E6D7}85.57 & \cellcolor[HTML]{D8E6D7}87.02 & \cellcolor[HTML]{D8E6D7}74.76 & \cellcolor[HTML]{D8E6D7}86.83 & \cellcolor[HTML]{D8E6D7}88.41 \\
\multirow{-2}{*}{Ours}            & Last                                  & 72.51                         & 84.74                         & 86.39                         & 74.12                         & 85.97                         & 87.76                         \\ \hline
\end{tabular}
\caption{Models performance comparison on test and validation sets using two different checkpoint choices: (a) \emph{Optimal:} A checkpoint selected based on the performance on the pixel-level annotated validation set, and (b) \emph{Last:} The checkpoint obtained after training each model for 20 epochs. 
 Please see \cref{sec_sup:checkpoint_choice} for the in-depth discussion.
 }
\label{tab:optimal_last_checkpoints}
\end{table*}

\subsection{Models Checkpoint Choice}
\label{sec_sup:checkpoint_choice}

As described in \cref{sec:implementation_details} of the main paper, for each of the models fine-tuned on sketch data: ours and competing methods, we select a checkpoint based on the performance on the validation set with pixel-level segmentation annotations, consisting of 475 sketches. 
This requires at training time having a small set of pixel-level annotated sketches, which can be limiting. 
However, we observe that the loss gradually decreases for our model, and it is safe to choose a last checkpoint if such an annotated set is not available.
In \cref{tab:optimal_last_checkpoints}, we provide a comparison with the results when for our model and competing models the last checkpoint is used. 
We trained for 20 epochs. 
We observe that after that the convergence rate is very low for each of the considered models. 

We observe only a marginal performance drop (less than one point in all metrics) for our model when the last checkpoint is used compared to a checkpoint selected based on the performance on the validation set (referred to as \emph{optimal} in the table). 
This implies that competitive model performance can be achieved without using any pixel-level annotations.

We also observe that with either of the choices of a checkpoint, the performance on the validation and test sets is similar, with just a small decrease in performance on the test set compared to the validation set.  
Our test set includes sketches from five non-expert artists whose sketches were not present in either the training or validation sets.
Therefore, this analysis implies that there is no over-fitting to the training data and our model robustly generalizes to the unseen sketches and drawing styles.

\begin{figure}[t]
  \centering
   \includegraphics[width=\linewidth]{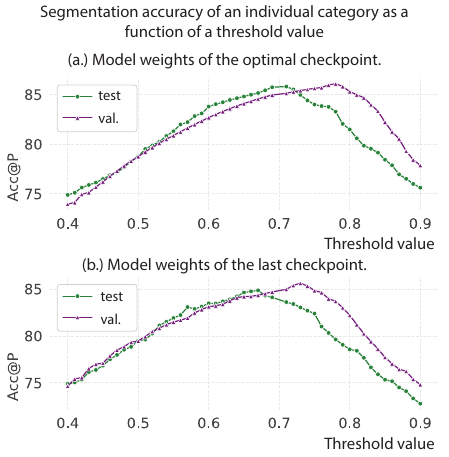} 
   \caption{\emph{Acc@P} values on test and validation sets (green and purple lines, respectively) for single category versus the rest segmentation task, as a function of a threshold value.
   The plots are shown for the two different choices of a checkpoint. (a) \emph{Optimal:} A checkpoint is selected based on the performance on the pixel-level annotated validation set, and (b) \emph{Last:} The checkpoint is obtained after training each model for 20 epochs. 
   Please see \cref{sec_sup:thresholding} for an in-depth discussion.}
   \label{fig:threshold_search}
\end{figure}

\begin{table*}[t]
% \vspace{-0.3cm}
\centering
\begin{tabular}{l|rrr|rr|rr}
 & \multicolumn{3}{c|}{Training} & \multicolumn{2}{c|}{Parameters} & \multicolumn{2}{c}{Inference} \\ \cline{2-8} 
 & \multicolumn{1}{l}{GFLOPS} & \multicolumn{1}{l}{Hours} & \multicolumn{1}{l|}{Epochs} & \# Trainable & \multicolumn{1}{l|}{\# Full} & \multicolumn{1}{l}{GFLOPS} & \multicolumn{1}{l}{Secs.} \\ \hline
CLIP        & 189.47 & 3.76 & 20    & 149M & 149M &  89.02& 0.21 \\
CLIPSurgery & 231.41 & 4.08 & 20    & 162M & 162M & 113.64 & 0.73 \\
Ours        & 346.36 & 7.31 & 20    & 16M& 165M & 121.59 & 1.05 \\ \hline
\end{tabular}%
% \vspace{-0.42cm}
\caption{\footnotesize{Note that the parameters of cross-attention layers (added complexity in our model over CLIPSurgery) are used only during training.
}}
% \vspace{-0.35cm}
\label{tab:cost}
\end{table*}

\subsection{Segmenting out Individual Categories}
\label{sec_sup:thresholding}
To explore the model's ability to isolate individual sketch categories through thresholding, as described in \cref{sec:inference} in the main paper, we assess two model versions, where (1) the \emph{optimal} checkpoint is used, selected based on the performance on the validation set and (2) the \emph{last} checkpoint is used (from the 20th epoch). 
We measure pixel accuracy (\emph{Acc@P}) of segmenting a sketch into an individual category and the rest (background), employing varying threshold values. 
\cref{fig:threshold_search} shows the plot of segmentation accuracy with different threshold values on test and validation sets when either optimal \cref{fig:threshold_search}(a.) or last \cref{fig:threshold_search}(b.) checkpoints are used. 

\paragraph{When optimal checkpoint is used} 
When using the optimal checkpoint, the model consistently achieves strong performance on validation and test sets, achieving $86.06\%$ and $85.71\%$ \emph{Acc@P}, respectively, albeit at different threshold values ($0.79$ and $0.71$, respectively). 
This implies that the label assignment confidence is slightly lower on the unseen sketches in new styles. 
However, despite this, the model maintains a consistently strong performance on these new sketches and styles.

\paragraph{When the last checkpoint is used} 
When we use the model from the last checkpoint, the best performance on the validation and test sets is obtained with slightly lower threshold values of $0.73$ and $0.68$, respectively. 
This implies that there is a correlation between the model's confidence and its performance.

\section{Computational Cost}
\label{sec_sup:Computational Cost}

We detail in \cref{tab:cost} the computational cost of our method compared to CLIP and CLIPSurgery.
Our two-level hierarchical network design introduces additional complexity, through value-value self-attention and cross-attention blocks. 
However, we maintain a comparable level of complexity to CLIPSurgery during inference. 
This slight computational increase is justified given the substantial 13 mIoU points improvement over CLIP\_Surgery\textsuperscript{$\star$$\star$} (as shown in \cref{tab:ablation_main} in the main paper). 
Our code can be further optimized to reach the performance of CLIPSurgery at inference time.

% \appendix
% \centering
% \onecolumn{
%  \Large{\textbf{Supplementary Material \\
% }\vspace{1.5em}
% }
% }
% % WARNING: do not forget to delete the supplementary pages from your submission 
% \input{sec_supplemental/supplemental}

\end{document}